\documentclass[10pt,twocolumn,letterpaper]{article}

\usepackage[pagenumbers]{iccv} 
\usepackage{graphicx}
\usepackage{booktabs}          
\usepackage{multirow}
\usepackage{amsmath}
\usepackage{bm}
\usepackage[numbers]{natbib}
\usepackage{float}
\usepackage{pifont} 
\usepackage{colortbl}  
\usepackage{array}

\definecolor{iccvblue}{rgb}{0.21,0.49,0.74}
\usepackage[pagebackref,breaklinks,colorlinks,allcolors=iccvblue]{hyperref}

\title{OmniSAM: Omnidirectional Segment Anything Model for UDA in Panoramic Semantic Segmentation}

\author{Ding Zhong$^{1,3,*}$\quad Xu Zheng$^{1,4,}$\thanks{Equal Contribution} \quad Chenfei Liao$^{1}$ \quad Yuanhuiyi Lyu$^{1}$ \quad Jialei Chen$^{5}$ \quad Shengyang Wu$^{3}$\\ Linfeng Zhang$^{6}$ \quad Xuming Hu$^{1,2,}$ \thanks{Corresponding author} \\ 
$^{1}$AI Thrust, HKUST(GZ) \quad $^{2}$CSE, HKUST \quad $^{3}$UMich \\ \quad $^{4}$INSAIT, Sofia University “St. Kliment Ohridski” \quad $^{5}$Nagoya University \quad $^{6}$SJTU \\
}

\begin{document}
\maketitle
\begin{abstract}

Segment Anything Model 2 (SAM2) has emerged as a strong base model in various pinhole imaging segmentation tasks. However, when applying it to $360^\circ$ domain, the significant field-of-view (FoV) gap between pinhole ($70^\circ \times 70^\circ$) and panoramic images ($180^\circ \times 360^\circ$) poses unique challenges. Two major concerns for this application includes 1) inevitable distortion and object deformation brought by the large FoV disparity between domains; 2) the lack of pixel-level semantic understanding that the original SAM2 cannot provide. To address these issues, we propose a novel \textbf{OmniSAM} framework, which makes the \textbf{first} attempt to apply SAM2 for panoramic semantic segmentation. Specifically, to bridge the first gap, OmniSAM first divides the panorama into sequences of patches. These patches are then treated as image sequences in similar manners as in video segmentation tasks. We then leverage the SAM2’s memory mechanism to extract cross-patch correspondences that embeds the cross-FoV dependencies, improving feature continuity and the prediction consistency along mask boundaries. For the second gap, OmniSAM fine-tunes the pretrained image encoder and reutilize the mask decoder for semantic prediction. An FoV-based prototypical adaptation module with dynamic pseudo label update mechanism is also introduced to facilitate the alignment of memory and backbone features, thereby improving model generalization ability across different sizes of source models. Extensive experimental results demonstrate that our method outperforms the state-of-the-art methods by large margins, \eg, 79.06\% (\textbf{10.22}\%$\uparrow$) on SPin8-to-SPan8, 62.46\% (\textbf{6.58}\%$\uparrow$) on CS13-to-DP13. The source code is available at \url{https://github.com/Ding-Zhong/OmniSAM}

\end{abstract}

\section{Introduction}
\label{sec:intro}

The expansive field of view (FoV) of $360^\circ \times 180^\circ$ brought by the panoramic cameras has led to their growing popularity in diverse applications, such as autonomous driving and virtual reality~\cite{elharrouss2021panoptic, ai2022deep, gao2022review, zheng2024open}. Unlike pinhole cameras, which are limited by a confined FoV, panoramic imaging enables a more holistic perception of the environment. Consequently, recent research efforts focus on leveraging this advantage to enhance scene understanding for intelligent systems.
Generally, panoramic cameras capture $360^\circ$ visual data in a spherical format, which is projected onto a 2D planar representation to facilitate compatibility with conventional imaging pipelines. Unlike pinhole images, this projection process introduces inevitable distortions and object deformations due to the non-uniform distribution of pixels. Moreover, developing effective panoramic segmentation models is challenging due to the scarcity of large-scale, precisely annotated datasets, as panoramic image annotation remains a labor-intensive task. For these reasons, unsupervised Domain Adaptation (UDA) for panoramic semantic segmentation~\cite{zhang2024behind,zheng2024360sfuda++,zheng2023look, zheng2023both, zheng2024semantics, jiang2025multi} has been proposed to bridge the domain gap between pinhole and the panoramic image domain.

Recently, the Segment Anything Model 2 (SAM2) achieves a breakthrough in both image and video instance segmentation, exhibiting remarkable zero-shot segmentation capabilities. This has generated significant interest in its potential applications across domains such as autonomous driving~~\cite{shan2023robustness, li2024fusionsam}, remote sensing~\cite{yan2023ringmo, liu2024pointsam, liu2024rsps}, and medical imaging~\cite{qin2024db, ma2024segment, fu2024cosam, wu2023medical}. 
In particular, recent methods have shown the effectiveness of SAM2 in semantic segmentation tasks, such as the Classwise-SAM-Adapter~\cite{pu2025classwise} for SAR image semantic segmentation and MLESAM~\cite{zhu2024customize} for multi-modal semantic segmentation, further emphasizing its strong adaptability. 
Intuitively, these successes raise the compelling question: ``\textit{Can SAM2 be effectively utilized in $360^\circ$ image semantic segmentation?}"

However, it is non-trivial to directly adapting SAM2 to the UDA for panoramic semantic segmentation, two significant problems persists, namely: \textbf{1)} the significant FoV gap, typically $70^\circ \times 70^\circ$ versus $180^\circ \times 360^\circ$,  between pinhole and panoramic domains, which complicates feature alignment across these distinct imaging modalities; \textbf{2)} the inherent limitation of SAM2, which provides instance-level masks but lacks the semantic knowledge required for precise semantic segmentation.
To address these challenges, we propose the \textbf{\textit{OmniSAM}} framework, which is the \textbf{first} attempt to accommodate SAM2 to panoramic semantic segmentation task.
The proposed \textbf{\textit{OmniSAM}} framework takes the SAM2 image encoder, which is fine-tuned through Low-Rank Adaptation (LoRA) \cite{hu2022lora}, as the backbone, and incorporates a customized semantic decoder for semantic segmentation. The FoV-based Prototypical Adaptation (FPA) method and dynamic pseudo label update strategy are applied for effective cross-domain feature alignment between pinhole and panoramic image sequences.

Specifically, we use the same processing pipeline for both  pinhole images (source domain) and panoramic images (target domain), where every raw image is divided into a \textit{\textbf{overlapping sequence of patches}} using a sliding window, which is significantly different from prior methods~\cite{zhang2024behind, zheng2024360sfuda++, zheng2023look, zheng2023both, zheng2024semantics, jiang2025multi}. The preprocessed sequences from pinhole and panoramic images are then used for source model training and target domain adaptation, respectively. As the image sequences resemble video frames, it can naturally leverage SAM2's memory mechanism. 
The memory encoder of SAM2 captures cross-patch correspondences within the input image sequences and encodes them into memory features. 
The memory attention (MA) module subsequently retrieves these stored features, enhancing cross-patch understanding. 

Given the assumption that the same patch/frame across different input image sequences exhibits shared distortions, object deformations, and statistical properties, our FPA method extracts each frame's prototype (\ie feature center) for alignment. The high-dimension feature representation consistency across domains is enforced by minimizing the Euclidean distance between these prototypes. 
The frame prototypes are updated iteratively and correspondingly at each adaptation step. Additionally, to mitigate the negative effect brought by the false pseudo-labels, we also introduce a dynamic pseudo-label update mechanism to ensure its prompt refinement to facilitate target domain self-supervised learning.

We conduct extensive experiments on the proposed OmniSAM framework and compare it with state-of-the-art (SoTA) methods \cite{zhang2024behind, zheng2023look, zheng2024360sfuda++} in Pinhole-to-Panoramic scenario. The results shows that our framework demonstrates superior performance over the SoTA methods by $\textbf{10.22}\%$ in indoor scenes and $\textbf{6.58}\%$ in outdoor scenes with trainable parameters less than $26$MB. In summary, our contributions can be summarized as: 1) We propose the \textbf{\textit{OmniSAM}} framework, making the \textbf{\textit{first}} attempt to explore the potential of SAM2 for the pinhole-to-panoramic semantic segmentation. 2) We introduce a dynamic pseudo label update strategy to mitigate the negative effects brought by fake pseudo labels during adaptation stage. 3) We propose a novel FPA module to effectively align features across domains and enhance the model's robustness against distortions in panoramic images.
\section{Related Works}
\label{sec:formatting}
\subsection{Segment Anything Model 2}
SAM2 \cite{ravi2024sam} is the second generation of the SAM \cite{kirillov2023segment}, designed for instance-level segmentation in both image and video tasks. Trained on \(50.9\)K videos with \(642.6\)K masklets, SAM2 features a powerful image encoder and memory-driven modules to capture cross-frame relations. Research on SAM focuses on improving instance-level segmentation \cite{yang2024samuraiadaptingsegmentmodel} and applying it to downstream tasks such as medical imaging \cite{ma2024segment, qin2024db} and remote sensing \cite{pu2025classwise}. 
Despite its ability to generate high-quality instance-level masks, SAM2 still lacks semantic knowledge. To address this, adapters and LoRA layers are introduced for domain fine-tuning \cite{chen2024sam2, chen2023sam,hu2022lora}, enabling knowledge transfer to improve performance in specific domains \cite{zhu2024customize, pu2025classwise}. In multi-modal semantic segmentation, several recent studies \cite{li2024fusionsamlatentspacedriven, xiao2024segmentmultiplemodalities, ma2024manetfinetuningsegmentmodel} have leveraged the pre-trained SAM/SAM2 backbone to extract high-dimensional representations from different modalities. These approaches introduce novel modules to fuse these features for optimal cross-domain representation. However, generalization to other domains still remains challenging.
In this paper, we apply SAM2 for UDA in panoramic semantic segmentation. We propose a OmniSAM framework, which adapts SAM2’s architecture and memory modules for enhanced knowledge transfer.

\begin{figure*}[ht!]
    \centering
    \includegraphics[width=1\textwidth]{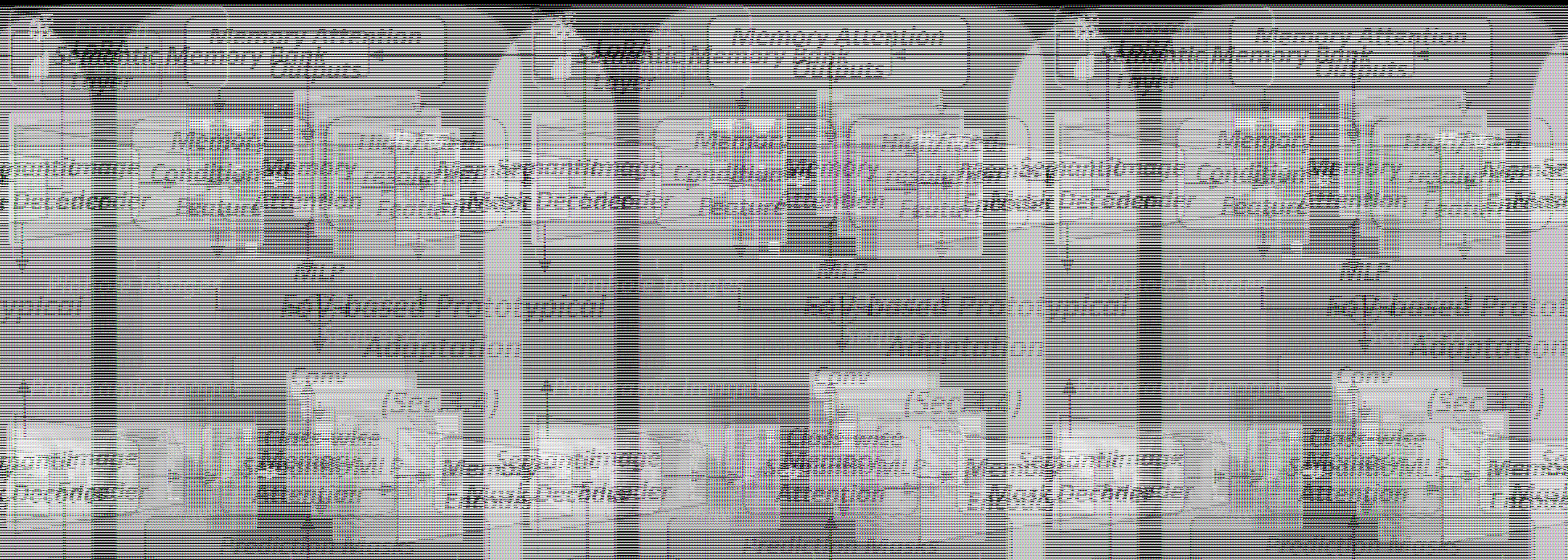}
    \caption{An overview of OmniSAM framework. First, OmniSAM is trained on the source domain to obtain the source model. Then, the FoV-based prototypical adaptation module is employed for cross-domain feature alignment.}
    \label{fig:Overview of OmniSAM}
\end{figure*}

\subsection{UDA for Panoramic Semantic Segmentation}
Domain adaptation enhances a model's generalization ability to unseen domains and has been widely applied in panoramic semantic segmentation \cite{zheng2023both,zheng2023look,zheng2024360sfuda++,zheng2024semantics,zhang2022bending,zhang2024behind,zhang2024goodsambridgingdomaincapacity,zhang2024goodsamplusbridgingdomaincapacity}. It can be categorized into three types: adversarial learning \cite{hoffman2018cycada, tsai2018learning, chang2019all}, pseudo-labeling \cite{zhang2024goodsambridgingdomaincapacity, zhang2024goodsamplusbridgingdomaincapacity}, and prototypical adaptation \cite{zheng2023both,zheng2024360sfuda++,zheng2024semantics,zheng2023look,zhang2022bending,zhang2024behind,zeng2024graph}. 
Adversarial training captures domain-invariant characteristics by using a discriminator to force the target model to generate indistinguishable features across domains. Pseudo-labeling generates self-supervised labels for the target domain. GoodSAM \cite{zhang2024goodsambridgingdomaincapacity} and GoodSAM++ \cite{zhang2024goodsamplusbridgingdomaincapacity} refine these labels using SAM, providing additional knowledge to the target model. 
The prototypical approach aligns high-level feature centers between domains. Previous methods \cite{zhang2022bending,zhang2024goodsambridgingdomaincapacity,zhang2024behind,zhang2024goodsamplusbridgingdomaincapacity,zheng2023both,zheng2023look,zheng2024360sfuda++,zheng2024semantics, li2023sgat4pass, hu2024deformable} focus on semantic details, and distortion in panoramic FoV during prototype extraction. However, these methodologies treat the equirectangular projection (ERP) images or tangent projection (TP) as pinhole images and forward the entire images to the model with no specific designs, causing significant prototypical gap across domains. 
To better accommodate the panoramas for SAM2 and efficient domain adaptation, we takes a totally different perspective on processing the target panoramas by generating \textbf{\textit{overlapping FoV sequences of patches}} with a sliding window cropping strategy. 

\section{Method}
\subsection{Overview}
As illustrated in Fig.~\ref{fig:Overview of OmniSAM}, we define the source domain and target domain as $S$ and $T$, respectively. Given a source pinhole image $x^S$ and a target panoramic image $x^T$, we apply a sliding window strategy to crop the image into overlapping sequences $\{x^S_1, x^S_2,\dots, x^S_t, \dots, x^S_N\}$ and $\{x^T_1, x^T_2,\dots, x^T_t, \dots x^T_N\}$, where $t$ denotes the patch index. Our model $F$ consists of two components: an image encoder $F_{en}$ and a customized mask decoder $F_{de}$. The encoder $F_{en}$ processes the input patch sequence sequentially, extracting multi-scale features at different resolutions—high, medium, and low—denoted as $f_{\text{high}}, f_{\text{med}},$ and $f_{\text{low}}$, respectively. If the memory attention is enabled, the low-resolution embedding $f_{\text{low}}$ is conditioned by past memory embeddings stored in a memory bank. The decoder $F_{de}$ then fuses these multi-scale features and generates the final predictions $\hat{y}^S$ or $\hat{y}^T$, taking either $\{f_{\text{high}}, f_{\text{med}}, f_{\text{low}}\}$ or $\{f_{\text{high}}, f_{\text{med}}, f_{\text{con}}\}$ as input. Additionally, a memory encoder projects the current patch output $\{f_{\text{low}}, \hat{y}\}$ into a memory embedding $g_t$ at $t$. To facilitate domain adaptation, we introduce an FoV-based Prototypical Adaptation module, enabling knowledge transfer between domains.

\subsection{SAM2 Adaptation}
Since SAM2 is designed for promptable image and video segmentation and requires external prompts such as bounding boxes or points, we introduce modifications to both the image encoder and the mask decoder to adapt the SAM2 backbone for the panoramic semantic segmentation tasks, enabling fixed-class prediction without the need for external prompts. \textbf{Image Encoder:} SAM2 \cite{ravi2024sam} leverages an MAE \cite{he2022masked} pretrained Hiera \cite{ryali2023hiera} image encoder, enabling the use of multiscale features during decoding. We apply LoRA fine-tuning to the query layers and value layers in the multi-scale attention block of the image encoder to extract semantic knowledge for our task, requiring fewer than 3MB of trainable parameters while maintaining efficiency and effectiveness. 
\textbf{Memory Mechanism:} For each processed frame, the memory encoder generates a memory embedding by downsampling the output mask with a convolutional module and adding it element-wise to the lowest-resolution (top-level) embeddings $f_\text{low}$ from the image encoder. This fused representation is then passed through additional convolutional layers for enhanced feature integration \cite{ravi2024sam}. The resulting memory embedding is stored in the model's memory bank, which has a fixed size of $n$. During each prediction step, memory attention conditions the current frame features using up to $n$ past memory embeddings. This is achieved by stacking $L$ transformer blocks, allowing the model to capture temporal dependencies and improve segmentation consistency across frames \cite{ravi2024sam}.
\textbf{Semantic Mask Decoder:} 
We propose a semantic decoder that leverages multi-scale embeddings $\{f_{\text{high}}, f_{\text{med}}, f_{\text{con}}\}$ for precise segmentation. First, linear projections are individually applied to these embeddings. Subsequently, embeddings $f_{\text{med}}$ and $f_{\text{con}}$ are spatially upscaled to match the resolution of $f_{\text{high}}$. The aligned embeddings are then fused via a convolutional module to produce a unified high-dimensional representation $\bm{f}$. Finally, a linear segmentation head is applied to $\bm{f}$ to obtain segmentation predictions.

\subsection{Dynamic Pseudo-label Updating Mechanism}
When working with large amounts of unlabeled data in the target domain, generating pseudo-labels for the entire dataset is computationally expensive and provides limited performance gains. As the model's accuracy improves in the target domain, periodically updating pseudo-labels becomes crucial to prevent prototype from over-fitting and ensure effective domain adaptation. To address this, we partition the adaptation process into multiple epochs, each consisting of fewer iterations. Within each epoch, a small subset of target domain samples is randomly selected for pseudo-label generation and subsequent adaptation. This strategy significantly enhances adaptation efficiency.

\begin{figure}[t!]
    \centering
    \includegraphics[width=0.47\textwidth]{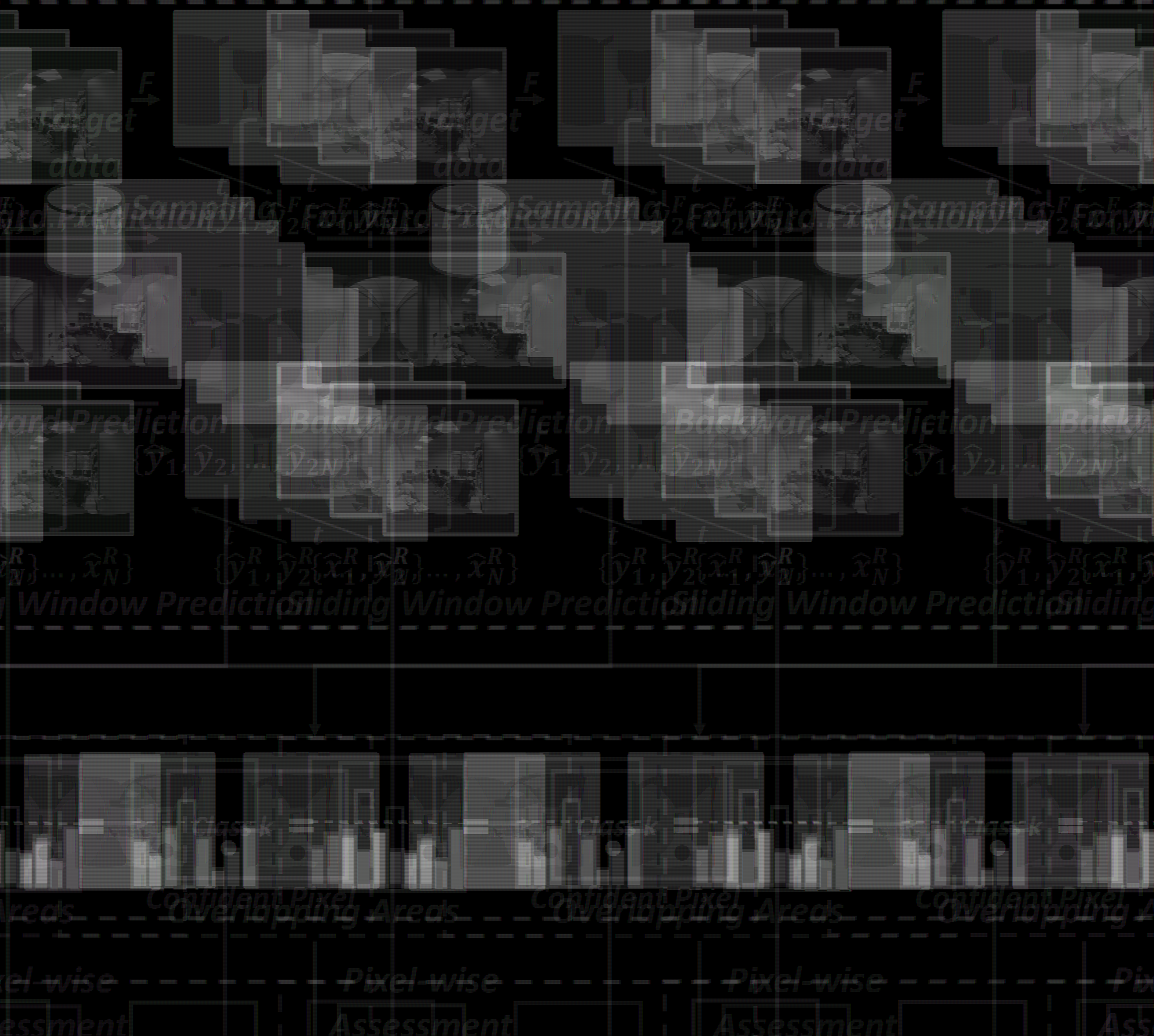}
    \caption{Dynamic pseudo-label updating mechanism.}
    \label{fig:Dynamic pseudo-label updating mechanism}
\end{figure}

As shown in Fig. \ref{fig:Dynamic pseudo-label updating mechanism}, the sampled image is patched into sequence and fed into the model in both forward and backward manners, denoted as $\{x_1^F, x_2^F, \dots, x_N^F\}$ and $\{x_1^R, x_2^R, \dots, x_N^R\}$, respectively. This bidirectional processing provides distinct memory contexts for each frame, resulting in different prediction maps. Then the corresponding output masks $\{\hat{y}_1, \hat{y}_2, \dots, \hat{y}_{2N}\}$ are aggregated for pixel-wise confidence assessment.

The second stage of Fig. \ref{fig:Dynamic pseudo-label updating mechanism} shows how the pixel-wise confidence assessment mechanism works. For each pixel in the original image, we aggregate all patch-level class predictions and logit values. \cite{zheng2024360sfuda++} proposed an approach that integrates prediction maps from different projections into the ERP image for pixel-wise assessment. Differently, our method evaluates overlapping regions \textbf{\textit{across prediction maps}}. The aforementioned bidirectional prediction processing further ensures that each pixel is covered by multiple patches, enhancing the assessment reliability.
In practical implementation, a coverage map is constructed to record how many patches overlap at each pixel. 
We then record, for each class, the total number of votes (\ie, how many overlapping patches predicted that class) as well as the minimum confidence among those patches. A pixel is assigned to a specific class only if \textit{1) all overlapping patches unanimously vote for that class and 2) the minimum confidence for that class surpasses a specified threshold.} Otherwise, the pixel is designated as an uncertain label. This strategy ensures that the final prediction at each pixel is both consistent across all overlapping patches and supported by sufficiently high confidence scores.

\begin{figure*}[t!]
    \centering
    \includegraphics[width=0.85\textwidth]{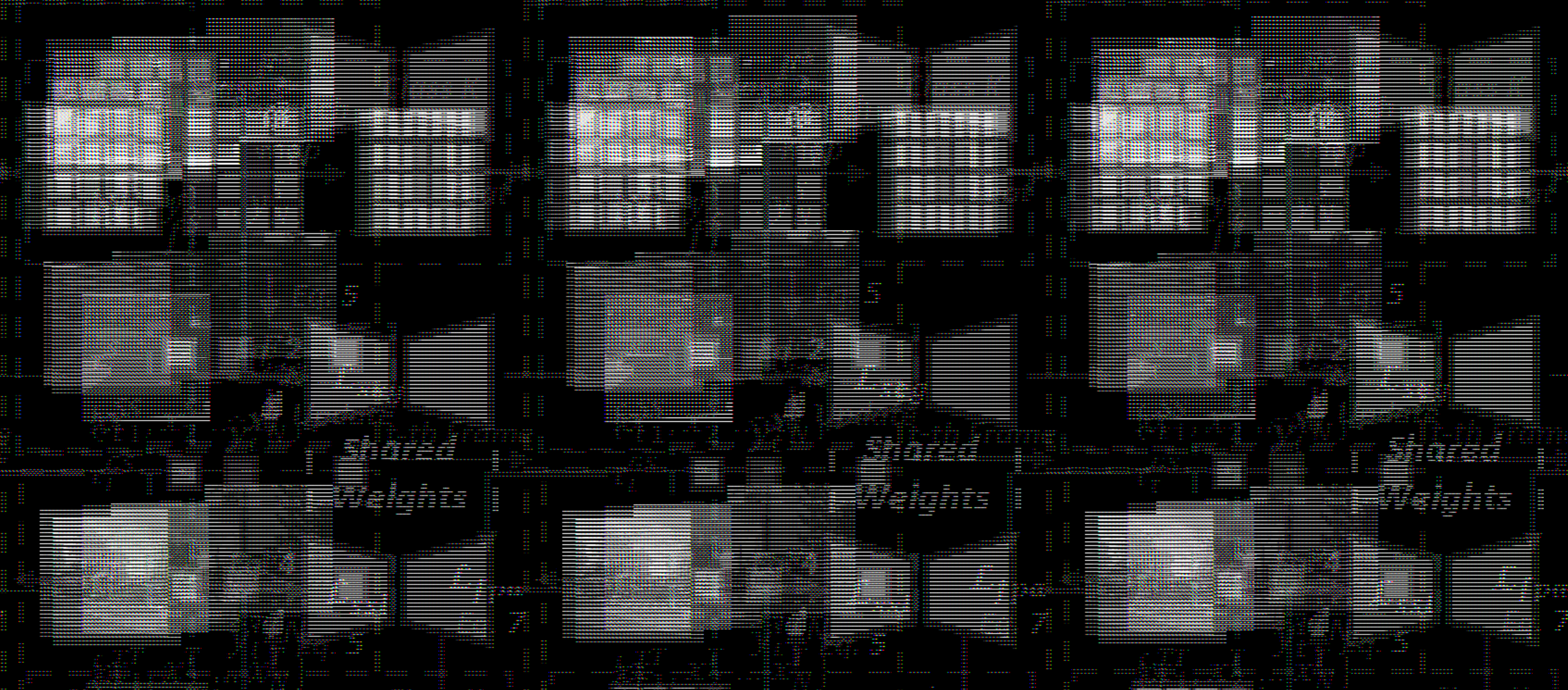}
    \caption{FoV-based Prototypical Adaptation.}
    \label{fig:FoV-based Prototypical Adaptation.}
\end{figure*}

\subsection{FoV-based Prototypical Adaptation (FPA)}
As preliminaries for domain adaptation, we first define the segmentation loss computed individually for the source and target domains. Given $\{x^S, y^S\}$ and $x^T$, we forward them into the pretrained source model and target model $F^T$ initialized with pretrained source weights:
\begin{equation}
\setlength{\abovedisplayskip}{3pt}
\setlength{\belowdisplayskip}{3pt}
    p^S = F_S(x^S), \quad p^T = F_T(x^T)
\end{equation}
where $p^S$ and $p^T$ are the corresponding prediction maps. The model's supervised loss is:
\begin{equation}
\setlength{\abovedisplayskip}{3pt}
\setlength{\belowdisplayskip}{3pt}
    \mathcal{L}_{seg} = -\sum_{k,i,j}^{K,H,W}y^S_{(k,i,j)}log(p^S_{(k,i,j)})
\end{equation}
This ensures that the model maintains its capabilities by learning class-specific features from a reliable ground-truth. Besides, we project the source model's predictions $P^T$ in the target domain into pseudo labels during the dynamic pseudo-label update stage:
\begin{equation}
\setlength{\abovedisplayskip}{3pt}
\setlength{\belowdisplayskip}{3pt}
\hat{y}^T_{(k, h, w)} = 1_{k \dot{=} \arg\max(p^T_{:,h,w})}
\end{equation}
where $k$ represents the class category index. Then self-supervised learning loss is used to optimize the model:
\begin{equation}
\setlength{\abovedisplayskip}{3pt}
\setlength{\belowdisplayskip}{3pt}
    \mathcal{L}_{ssl} = -\sum_{k,i,j}^{K,H,W}\hat{y}^T_{(k,i,j)}log(p^T_{(k,i,j)})
\end{equation}

To effectively transfer knowledge to the target domain, we propose an FoV-based Prototypical Adaptation (FPA) module, as shown in Fig. \ref{fig:FoV-based Prototypical Adaptation.}. Specifically, on the decoder side of our model, the multi-scale feature maps are concatenated separately and fused using a convolutional module, as shown in Fig. \ref{fig:Overview of OmniSAM}. We denote the resulting feature maps as \(\bm{f^S}, \bm{f^T} \in \mathbb{R}^{C \times H \times W}\), where \(C\) is the dimension of the feature, and \(H\) and \(W\) are the height and width of the feature maps. 
Given the associated labels or pseudolabels, we first resize these labels to the same size as the aforementioned feature maps. Then the source prototype and target prototype \(\tau^S, \tau^T \in \mathbb{R}^{M\times K \times C}\) at the FoV-sequence prediction step $t$
are computed as:
\begin{equation}
\setlength{\abovedisplayskip}{3pt}
\setlength{\belowdisplayskip}{3pt}
    \tau^k_t=\frac{1}{M}\sum^{H,W}_{i,j}(y_{(k,i,j)})_t\cdot(\bm{f}_{(i,j)})_t,\quad0\le t \le N-1
\end{equation}
where \(\tau^k\in\mathbb{R}^{C}\) denotes the prototype of class \(k\), \(M\) is the element-wise sum of the binary map \(y^k\) and $N$ is the size of memory bank.
During training, the global source prototype is iteratively updated using the batch source prototype through the following equation:
\begin{equation}
\setlength{\abovedisplayskip}{3pt}
\setlength{\belowdisplayskip}{3pt}
    (\tau^{GS}_{t})_n = (1-\frac{1}{n}) (\tau^{GS}_{t})_{n-1} + \frac{1}{n} (\tau^S_{t})_n
\end{equation}
where \(n\) is the iteration number and $t$ represents the frame index. In particular, we use the \textit{frozen source model} for the computation of source prototype. After this update, the target prototype of frame $t$ is aligned with its global source prototype using the Mean Squared Error (MSE) loss to bridge the class-wise knowledge gap between domains:
\begin{equation}
\setlength{\abovedisplayskip}{3pt}
\setlength{\belowdisplayskip}{3pt}
    \mathcal{L}_{fpa} = \frac{1}{KC} \| \tau^{GS}_{t} - \tau^{T}_{t} \|_F
\end{equation}
where \(\|\cdot\|_F\) represents the Frobenius norm. This drives alignment between the source feature \(\bm{f^S}\) and the target feature \(\bm{f^T}\).
Thus, the training objective for learning the target model is defined as:
\begin{equation}
\setlength{\abovedisplayskip}{3pt}
\setlength{\belowdisplayskip}{3pt}
    \mathcal{L} = \mathcal{L}_{seg}+\mathcal{L}_{ssl}+\lambda \mathcal{L}_{fpa}
\end{equation}
where $\mathcal{L}^S_{seg}$ is the supervised segmentation loss in source domain, $\mathcal{L}^T_{ssl}$ refers to the self-supervised segmentation loss in target domain based on pseudo-labels, $\mathcal{L}_{fpa}$ denotes the MSE loss from FPA and $\lambda$ is the super-parameter. 

Our FPA approach differs from previous UDA methods for panoramic segmentation in two key aspects:
\textit{\textbf{(a)}} It samples only \textbf{\textit{a small subset}} of the target dataset for pseudo-labeling and adaptation in each epoch. This ensures that as the model improves in the target domain, pseudo-labels are promptly updated, minimizing the impact of incorrect labels.
\textbf{(b)} It uses a frozen pretrained source model to \textbf{\textit{iteratively aggregate class-wise prototypes}} from the source domain. These prototypes serve as reliable references for adaptation, guiding the alignment of target prototypes with the source feature space.
\begin{table}[t!]
    \centering
        \small 
        \setlength{\tabcolsep}{6pt}


        \resizebox{0.45\textwidth}{!}{  
        \begin{tabular}{lccc}
        \toprule
        \multicolumn{4}{l}{\textbf{(1) Indoor Pin2Pan:}} \\
        \midrule
        {Network} & {SPin8} & {SPan8} & {mIoU Gaps} \\
        \midrule
        Trans4PASS+-S~\cite{zhang2024behind} & 67.28 & 63.73 & -3.55 \\
        OmniSAM-T    & 66.20 & 66.70 {(+2.97)} & +0.50 \\
        OmniSAM-S    & 68.76 & 69.56 {(+5.83)} & +0.80 \\
        OmniSAM-B    & 72.65 & 72.32 {(+8.59)} & -0.33 \\
        OmniSAM-L    & 75.18 & 76.85 {(\textbf{+13.12})} & +1.67 \\
        \midrule
        \multicolumn{4}{l}{\textbf{(2) Outdoor Pin2Pan:}} \\
        \midrule
        {Network} & {CS13} & {DP13} & {mIoU Gaps} \\
        \midrule
        Trans4PASS+-S~\cite{zhang2024behind} & 74.52 & 51.48                   & -23.04 \\
        OmniSAM-T    & 76.50 & 54.99 {(+3.59)} & -22.51 \\
        OmniSAM-S    & 77.13 & 56.12 {(+4.64)} & -21.52 \\
        OmniSAM-B    & 77.00 & 56.49 {(+5.01)} & -20.51 \\
        OmniSAM-L    & 79.50 & 56.61 {(\textbf{+5.13})} & -22.89 \\
        \midrule
        \multicolumn{4}{l}{\textbf{(3) Outdoor Syn2Real:}} \\
        \midrule
        {Backbone} & {SP13} & {DP13} & {mIoU Gaps} \\
        \midrule
        Trans4PASS+-S~\cite{zhang2024behind} & 61.59 & 43.17 & -18.42 \\
        OmniSAM-B    & 67.06 & 45.51 {(\textbf{+2.34})} & -22.32 \\
        \bottomrule
        \end{tabular}
        }

        \caption{Pin2Pan vs. Syn2Real domain gaps.}
        \label{tab:Pin2Pan domain gaps}
\end{table}

\begin{table*}[ht]
    \centering
    \setlength{\tabcolsep}{10pt}
    \resizebox{\textwidth}{!}{
    \begin{tabular}{llccccccccccccccc}
    \toprule
    {Method} & {Network} & {mIoU} & {Ceiling} & {Chair} & {Door} & {Floor} & {Sofa} & {Table} & {Wall} & {Window} & $\Delta$\\
    
    \midrule
    
    \multirow{3}{*}{MPA} 
    & PVT-S \cite{wang2021pyramid} & 57.95 & 85.85 & 51.76 & 18.39 & 90.78 & 35.93 & 65.43 & 75.00 & 40.43 & - \\
    & Trans4PASS-S~\cite{zhang2022bending} & 64.52 & 85.08 & 58.72 & 34.97 & 91.12 & 46.25 & 71.72 & 77.58 & 50.75 & - \\
    & Trans4PASS+-S~\cite{zhang2024behind}  & 67.16 & 90.04 & 64.04 & 42.89 & 91.74 & 38.34 & 71.45 & 81.24 & 57.54 & - \\
    \midrule
    \multirow{1}{*}{SFUDA}
    & 360SFUDA++ w/ b2 ~\cite{zheng2024360sfuda++}& 68.84 &85.50 & 57.59 & 53.15 & 87.40 & 53.63 & 66.49 & 80.23 & 66.75 & * \\ 
    \midrule
    \multirow{8}{*}{Ours} 
    & OmniSAM-T w/o MA& 68.72 & 92.12 & 64.62 & 35.54 & 94.21 & 37.70 & 74.33 & 81.70 & 69.50 & -0.12 \\
    & OmniSAM-S w/o MA& 70.65 & 91.46 & 65.41 & 55.10 & 94.56 & 33.88 & 75.61 & 84.53 & 64.66 & +1.81 \\ 
    & OmniSAM-B w/o MA& 73.09 & 91.63 & 69.33 & 60.43 & 94.45 & 36.16 & 76.22 & 85.26 & 71.26 & +4.25 \\ 
    & OmniSAM-L w/o MA&  78.02 & \textbf{93.79} &  71.19 & \textbf{78.07} & \textbf{95.17} &  47.25 &  81.77 & \textbf{89.85} & 66.51 &  +9.18 \\ \cmidrule{2-12}
    & OmniSAM-T w/ MA& 69.10 & 92.10 & 64.60 & 36.97 & 94.25 & 38.86 & 74.16 & 81.78 & 70.09 & +0.26 \\
    & OmniSAM-S w/ MA& 70.81 & 91.74 & 66.46 & 66.74 & 94.88 & 12.35 & 77.43 & 86.90 & 69.95 & +1.97 \\
    & OmniSAM-B w/ MA& 74.72 & 91.06 & 66.65 & 69.31 & 94.57 & 36.79 & 76.98 & 86.58 & \textbf{75.87} & +5.88 \\ 
    & OmniSAM-L w/ MA&  \textbf{79.06} & 93.25 & \textbf{72.12} & 77.97 & 95.00 & \textbf{52.08} & \textbf{81.82} &  89.62 & 70.58 & \textbf{+10.22} \\
    \bottomrule
    \end{tabular}
    }
    \caption{Per-class results of the Stanford2D3D pinhole-to-panoramic scenario (* denotes the baseline).}
    \label{tab:Per-class results of the Stanford2D3D pinhole-to-panoramic scenario}
\end{table*}

\section{Experiments}
\subsection{Experimental Setup}
\noindent \textbf{Datasets}: To evaluate our proposed OmniSAM model and UDA framework, we conducted experiments on five datasets including two real-world pin2pan scenarios and one syn2real scenario. \textit{\textbf{Stanford2D3D-Pinhole}}: The Stanford2D3D Pinhole (SPin) dataset~\cite{armeni2017joint} consists of 70,496 pinhole images with annotations across 13 categories. \textit{\textbf{Stanford2D3D-Panoramic}}: The Stanford2D3D Panoramic (SPan) dataset~\cite{armeni2017joint} contains 1,413 panoramic images, sharing the same 13 semantic categories with the SPin dataset.
\textit{\textbf{Cityscapes}}: The Cityscapes (CS) dataset~\cite{cordts2016cityscapes} has 2,979 training images and 500 validation images, all annotated with 19 semantic classes. 
\textit{\textbf{SynPASS}}: The SynPASS (SP)~\cite{zhang2024behind} dataset is a synthetic dataset composed of 9080 synthetic panoramic images annotated across 22 categories. The sets for training, validation, and testing contain 5700, 1690, and 1690 images, respectively. \textit{\textbf{DensePASS}}: The DensePASS (DP) dataset~\cite{ma2021densepass} consists of 2,000 unlabeled panoramic images and 100 labeled images for final testing, sharing the same 19 semantic classes as Cityscapes.

\noindent\textbf{UDA settings}: To keep the same setting with \cite{zhang2024behind} and \cite{zheng2024360sfuda++}, we focus on a subset of 8 categories for the SPin and SPan datasets, and 13 categories for the CS, SP and DP datasets. There are three scenarios being investigated: (1) Indoor Pin2Pan: \textit{\textbf{SPin8-to-SPan8}}. (2) Outdoor Pin2Pan: \textit{\textbf{CS13-to-DP13}}. (3) Outdoor Syn2Real: \textit{\textbf{SP13-to-DP13}}.

\noindent \textbf{Implementation Details:}
Since the SAM2 backbone only accepts square-shaped images, we design two processing pipelines for source model training to ensure generalization across different cases: one for rectangular inputs and another for square-shaped inputs, as shown in \textbf{\textit{suppl. mat.}}.

\noindent \textbf{Preprocessing:} All models are trained on 2 NVIDIA A800 GPU with an initial learning rate of \(6\times10^{-5}\), scheduled using a polynomial decay strategy with a power of \(0.9\). We use the AdamW optimizer with epsilon of \(10^{-8}\) and weight decay of \(10^{-4}\). The batch size is set to 4 for OmniSAM-L and OmniSAM-B, and 6 for OmniSAM-S and OmniSAM-T per GPU. The memory bank size is set to 9. We use a sliding window to crop the rectangular image, resulting in a 9-frame FoV sequence of \(1024\times1024\) images. The sliding stride depends on the resolution of the images. The SPin dataset provides \(1080\times1080\) images, which are resized to \(1024\times1024\) and do not require sliding. Meanwhile, the stride is set to \(128\) for both the CS and SP datasets with \(2048\times1024\) images. The SPan dataset contains \(4096\times2048\) images, which include inherent black regions brought by equirectangular projection. Hence we remove these redundant areas by cropping the images and downscale them to \(3072\times1024\). The corresponding sliding stride is \(256\). The DensePASS dataset contains \(2048\times400\) images, which are resized to \(4096\times1024\), with a sliding stride of \(384\).

\noindent \textbf{Model Parameters:}
The total parameters and computation costs of our OmniSAM variants are presented in the \textit{\textbf{suppl. mat.}}. The \textit{\textbf{Tiny}} and \textit{\textbf{Small}} variants exhibit competitive parameter counts, similar to state-of-the-art methods. While the \textit{\textbf{Large}} (209.2M) and the \textit{\textbf{Base}} (72.0M) variants significantly increase the model size, which may limit its real-time application. A trade-off in inference speed should be considered while choosing the model for specific task.

\begin{table*}[ht]
    \centering
    \setlength{\tabcolsep}{4pt}

    \normalsize
    
    \resizebox{\textwidth}{!}{
    \begin{tabular}{llccccccccccccccc}
    \toprule
    {Method} & {Network} & {mIoU} & {Road} & {S.Walk} & {Build.} & {Wall} & {Fence} & {Pole} & {Tr.L} & {Tr.S} & {Veget.} & {Terrain} & {Sky} & {Person} & {Car} & $\Delta$\\
    \midrule
    
    \multirow{1}{*}{MPA} 
    & Trans4PASS+-S~\cite{zhang2024behind}  & 55.24 & 82.25 & 54.74 & 85.80 & 31.55 & 47.24 & 31.44 & 21.95 & 17.45 & 79.05 & 45.07 & 93.42 & 50.12 & 78.04 & - \\
    \multirow{1}{*}{CFA} 
    & DATR-S~\cite{zheng2023look}   & 55.88 & 80.63 & 51.77 & 87.80 & \textbf{44.94} & 43.73 & 37.23 &  25.66 &  21.00 & 78.61 & 26.68 & 93.77 & 54.62 & 80.03  & *\\
    \multirow{1}{*}{MSDA} 
    & DTA4PASS~\cite{jiang2025multi}  & 57.16 & 80.35 & 53.24 & 87.93 & 32.46 & 48.03 & 30.97 & 27.47 & 19.32 & 80.40 & \textbf{50.06} & 94.34 & 56.31 & 82.18 & -  \\
    \midrule
    
    \multirow{8}{*}{Ours} 
    & OmniSAM-T w/o MA& 53.73 & 79.03 & 42.19 & 86.09 & 28.28 & 45.95 & 35.19 & 10.20 & 20.53 & 79.41 & 35.49 & 94.40 & 63.81 & 77.88 & -2.15  \\ 
    & OmniSAM-S w/o MA& 57.03 & 78.99 & 49.57 & 88.77 & 38.48 & 47.47 & 38.77 & 21.84 & 15.81 & 81.12 & 39.32 & 94.72 & 65.36 & 81.22 & +1.15  \\
    & OmniSAM-B w/o MA& 59.34 & 81.69 & 53.87 & 89.33 & 39.74 & 50.84 & 41.98 & 20.54 & 21.50 & 81.71 & 44.63 & 95.06 & 68.13 & 82.34 & +3.46  \\
    & OmniSAM-L w/o MA & 59.02 & 83.49 & 56.14 & 88.29 & 34.29 & 52.39 & 38.81 & 23.97 & 19.83 & \textbf{82.52} & 44.84 & \textbf{95.26} & 61.97 & \textbf{85.43} & +3.14  \\ \cmidrule{2-17}
    & OmniSAM-T w/ MA& 59.01 & 79.95 & 45.78 & 88.03 & 39.74 & 47.99 & 42.68 & 26.69 & 29.55 & 78.00 & 42.98 & 94.56 & 68.23 & 82.98 & +3.13  \\
    & OmniSAM-S w/ MA& 60.23 & 80.76 & 46.36 & 89.79 &  44.46 & 48.68 &  \textbf{45.32}&   29.33 & 25.15 & 79.51 &  46.28 & 94.41 & 68.90 & 84.06 & +4.35  \\
    & OmniSAM-B w/ MA&  \textbf{62.46} &  \textbf{84.02} &  \textbf{56.23} &  89.93 & 44.01 &  54.54 & 44.50 & 25.19 &  \textbf{33.42} &  81.77 &  49.16 & 94.69 &  \textbf{71.64} &  82.89 &  \textbf{+6.58}  \\  
    & OmniSAM-L w/ MA&  61.63 &  82.45 &  53.65 & \textbf{90.05} & 44.00 &  \textbf{54.75} & 43.36 &  \textbf{30.99} & 28.27 &  80.04 & 43.59 & 94.48 &  70.70 &  84.88 &  +5.75  \\
    \bottomrule
    \end{tabular}
    }
    \caption{Per-class results of the Cityscapes13-to-DensePASS13 scenario (* denotes the baseline).}
    \label{tab:Per-class results of the Cityscapes13-to-DensePASS13 scenario}
\end{table*}

\begin{table*}[ht]
    \centering
    \setlength{\tabcolsep}{4pt}
    \normalsize
    \resizebox{\textwidth}{!}{
    \begin{tabular}{llcccccccccccccc}
    \toprule
    Method & Network & mIoU & Road & S.Walk & Build. & {Wall} & {Fence} & {Pole} & {Tr.L} & {Tr.S} & {Veget.} & {Terrain} & {Sky} & {Person} & {Car} \\
    \midrule
    
    \multirow{5}{*}{Source-only} 
    & SegFormer-B1 \cite{xie2021segformer} & 35.81 & 63.36 & 24.09 & 80.13 & 15.68 & 13.39 & 16.26 & 7.42 & 0.09 & 62.45 & 20.20 & 86.05 & 23.02 & 53.37 \\
    & Trans4PASS+-S \cite{zhang2024behind}  & 43.17 & 73.72 & 43.31 & 79.88 & 19.29 & 16.07 & 20.02 &  8.83 &  1.72 & 67.84 & 31.06 & 86.05 & 44.77 & 68.58 \\
    & DATR-S \cite{zheng2023look}  & 35.29 & 60.43 & 13.57 & 76.69 & 18.35 & 5.88 & 17.33 &  3.44 &  2.62 & 62.68 & 19.54 & 83.58 & 34.30 & 58.56 \\ \cmidrule{2-16}
    & OmniSAM-B w/o MA& 41.53 & 59.54 & 10.57 & 79.39 & 33.35 & 25.16 & 25.47 & 11.54 & 9.29 & 72.31 & 18.05 & 90.08 & 42.37 & 62.74 \\
    & OmniSAM-B w/ MA  & 45.51 & 66.24 & 17.45 & 81.57 & 34.23 & 26.17 & 29.11 & 17.39 & 12.67 & 71.29 & 18.45 & 89.41 & 53.51 & 74.10 \\
    \midrule
    
    \multirow{1}{*}{SSL} 
    & PVT \cite{wang2021pyramid} & 38.74 & 55.39 & 36.87 & 80.84 & 19.72 & 15.18 & 8.04 & 5.39 & 2.17 & 72.91 & 32.01 & 90.81 & 26.76 & 57.40 \\ 
    \midrule
    
    \multirow{4}{*}{MPA} 
    & PVT \cite{wang2021pyramid} & 40.90 & 70.78 & 42.47 & 82.13 & 22.79 & 10.74 & 13.54 & 1.27 & 0.30 & 71.15 & 33.03 & 89.69 & 29.07 & 64.73 \\ 
    & Trans4PASS-T \cite{zhang2022bending}  & 45.29 & 67.28 & 43.48 & 83.18 & 22.02 & 21.98 & 22.72 & 7.86 & 1.52 & 73.12 & 40.65 & 91.36 & 42.69 & 70.87 \\
    & Trans4PASS+-S \cite{zhang2024behind}  & 50.88 & 77.74 & 51.39 & 82.53 & 29.33 & 43.37 & 25.18 & 20.09 &  8.37 & 76.36 & 41.56 & 91.07 & 45.43 & 68.98 \\
    & DATR-S \cite{zheng2023look} & 52.76 & 78.33 & \textbf{52.70} & 85.15 & 30.69 & 42.59 & 32.19 & 24.20 & 17.90 & 77.72 & 27.24 & 93.86 & 47.98 & 75.34 \\
    \midrule

    \multirow{1}{*}{CFA} 
    & DATR-S \cite{zheng2023look} & 54.05 & \textbf{79.07} & 52.28 & 85.98 & 33.38 & \textbf{45.02} & 34.47 &  \textbf{26.15} &  18.27 &\textbf{78.21} & 26.99 & \textbf{94.02} & 51.21 & 77.62 \\
    \midrule
    
    \multirow{1}{*}{Ours} 
    & OmniSAM-B w/ MA& \textbf{54.61} & 75.49 & 27.87 & \textbf{88.48} & \textbf{43.86} & 37.89 & \textbf{35.98} & 18.75 & \textbf{ 22.67} & 76.35 & \textbf{43.84} & 92.89 & \textbf{65.57} & \textbf{80.27} \\  
    \bottomrule
    \end{tabular}
    }
    \caption{Per-class results of the SynPASS13-to-DensePASS13 scenario.}
    \label{Per-class results of the SynPASS13-to-DensePASS13 scenario}
\end{table*}

\subsection{Source Model Domain Gaps}

We quantify the Pin2Pan domain gaps by comparing performance between standard pinhole (SPin8/CS13) and panoramic (SPan8/DP13) settings in both indoor and outdoor scenarios. As shown in Table \ref{tab:Pin2Pan domain gaps}, OmniSAM models consistently outperform the baseline.
In indoor scenes, the OmniSAM models demonstrate strong generalization capabilities without requiring adaptation techniques. 
Notably, OmniSAM-L surpasses Trans4PASS+-S by over \(13\%\) on indoor panoramic images, achieving \(73.77\%\), indicating superior adaptation to panoramias. The \textit{tiny}, \textit{small}, and \textit{large} variants of OmniSAM achieves even better performance on the target domain than on the source domain. These results may seem counterintuitive, as the model's performance in the target domain even surpasses that in the source domain. This is very likely the results of the low-quality ground truth annotations of small objects (\textit{e.g. chair, sofa, table}) in the target domain test set. Nevertheless, the significant improvement on this benchmark and visualizations still demonstrate the superiority of our model. 
For outdoor Pin2Pan test case, the Pin2Pan gaps are more pronounced, with Trans4PASS+-S experiencing a significant decline of \(-23.04\%\) in mIoU, highlighting its challenge in handling panoramic distortions. In contrast, the OmniSAM models consistently reduce this gap, with OmniSAM-B showing the smallest drop of \(-20.51\%\) in mIoU. 
Despite the domain gap, OmniSAM achieves superior performance in the target domain, with mIoU improvements ranging from \(3.59\%\) to \(5.21\%\). 
In outdoor Syn2Real scenario, OmniSAM-B outperforms Trans4PASS+-S by $2.34\%$.

\begin{figure}[t!]
    \centering
    \includegraphics[width=0.47\textwidth]{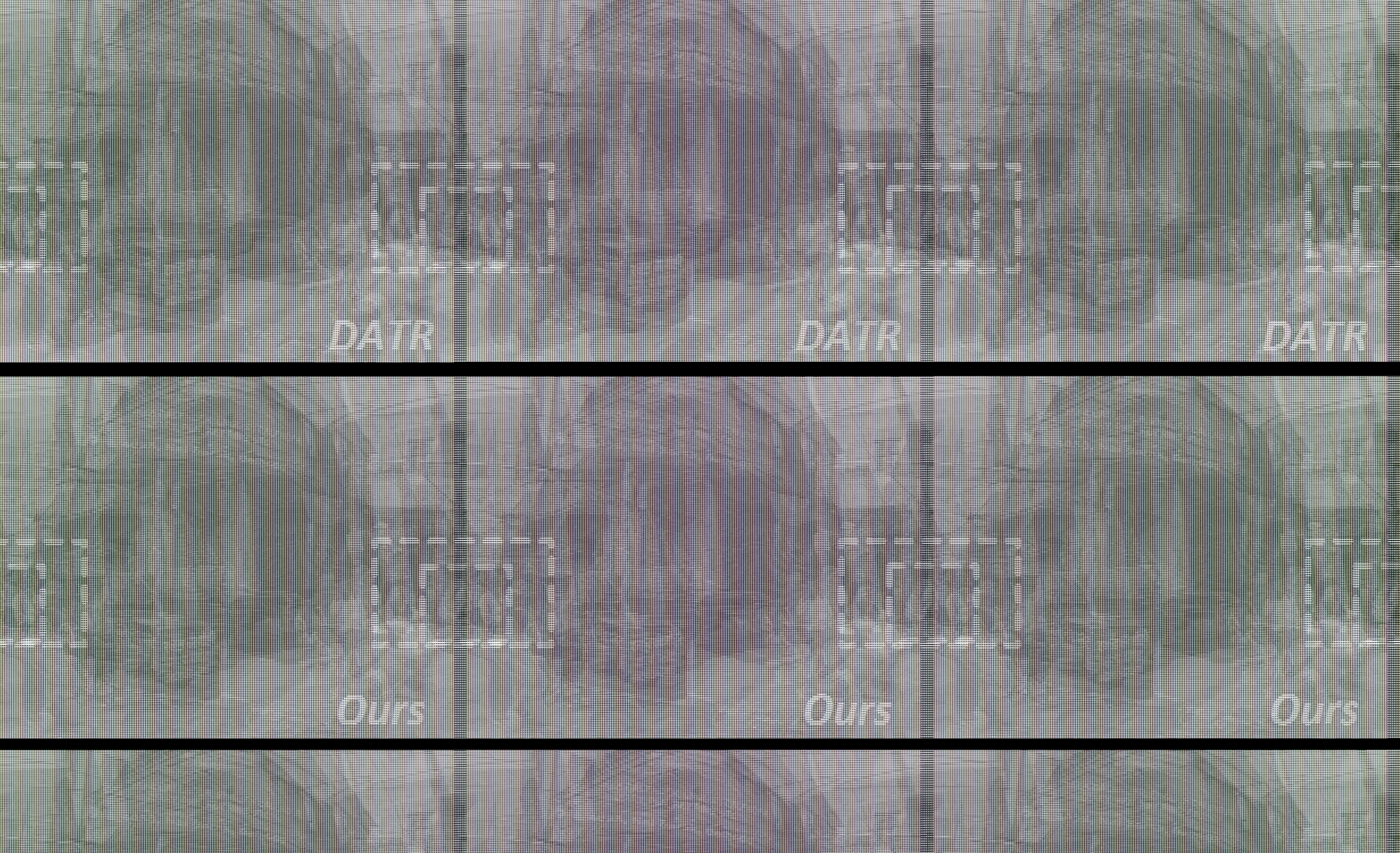}
    \caption{Visualizations on DensePASS dataset.}
    \label{fig:Visualizations_dp}
\end{figure}

\begin{figure}[t!]
    \centering
    \includegraphics[width=0.47\textwidth]{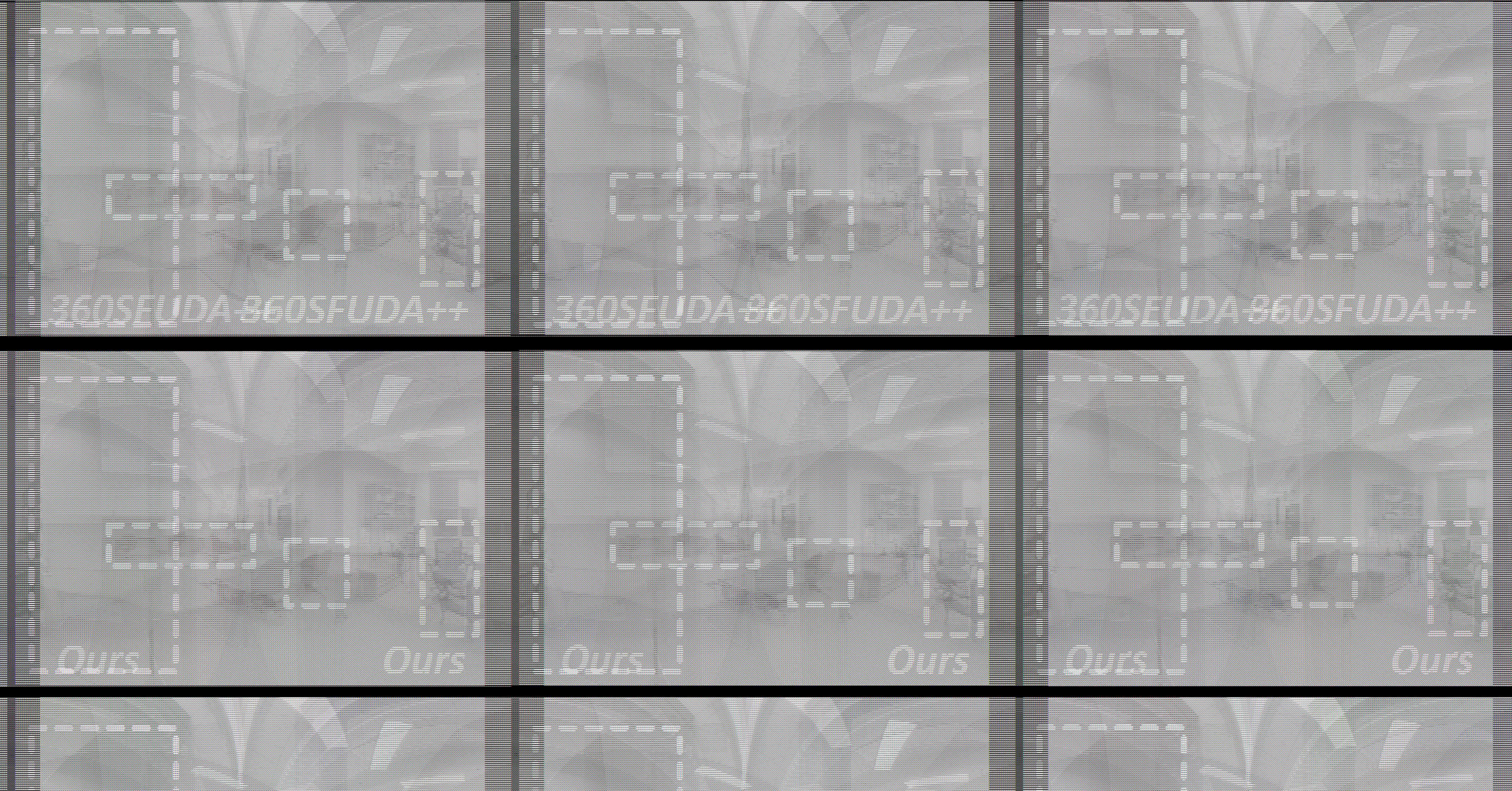}
    \caption{Visualizations on Stanford2D3D-Panoramic dataset.}
    \label{fig:Visualizations_span}
\end{figure}
\subsection{Experimental Results}

First, we evaluate our proposed framework under the SPin8-to-SPan8 setting. The results for source model are presented in Fig. \ref{fig:Visualizations_span} and Table \ref{tab:Per-class results of the Stanford2D3D pinhole-to-panoramic scenario}, demonstrating both the source model capabilities and the effectiveness of adaptation. Given the source model performance in Table \ref{tab:Pin2Pan domain gaps}, the \textit{small}, \textit{base}, \textit{large} variants of OmniSAM outperform the baseline model, 360SFUDA++, even without requiring adaptation techniques. Notably, the large OmniSAM variant achieves the highest mIoU score of \(76.85\%\), demonstrating superior generalization in the indoor Pin2Pan task. 
Subsequently, we test our FPA method on both OmniSAM with memory attention w/ and w/o MA. The results indicate significant improvements on each model after bridging the knowledge gap. Our framework outstrips 360SFUDA++ by a margin of $+0.26\%$ to $+10.22\%$ in mIoU. Notably, OmniSAM-L with memory modules scores the highest $79.06\%$, achieving the SoTA performance on the indoor Pin2Pan task. 

Next, we evaluate our framework in the CS13-to-DP13 setting. As shown in Table \ref{tab:Per-class results of the Cityscapes13-to-DensePASS13 scenario}, with adaptation incorporated, all variants surpass the DTA4PASS model \cite{jiang2025multi}, which is based on multiple source datasets, achieving improvement in mIoU of \(3.13\%\) to \(6.58\%\). Even without memory modules, OmniSAM-S, OmniSAM-B and OmniSAM-L outperform the baseline DATR-S, which employs CFA adaptation \cite{zheng2023look}. The visualization results in Fig. \ref{fig:Visualizations_dp} also demonstrate the superiorty of our OmniSAM.
Additionally, we explore the capability of our model within the Syn2Real scenario. The quantitative results of SP13-to-DP13 is presented in Table \ref{Per-class results of the SynPASS13-to-DensePASS13 scenario}. The OmniSAM-B yields $0.56\%$ mIoU improvement. More quantitative comparison and visualization results refer to the \textbf{\textit{suppl. mat.}}.
\section{Ablation Study}
\begin{figure}[t!]
    \centering
    \includegraphics[width=0.47\textwidth]{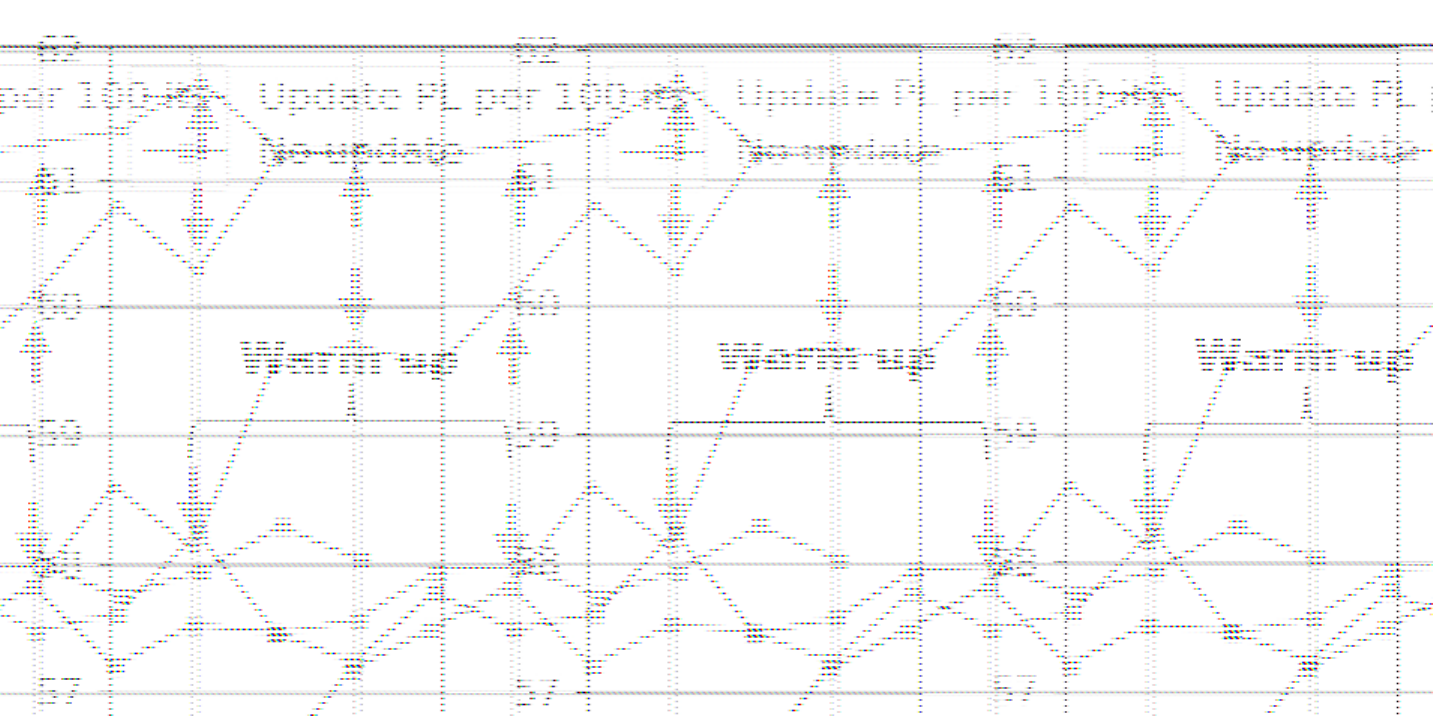}
    \vspace{-12pt}
    \caption{Ablation study of the pseudo-label update mechanism (red arrows indicate update points). A warm-up is performed at first for global prototypes computation.} 
    \label{fig:Ablation of pseudo-label update mechanism.}
\end{figure}

\noindent \textbf{Ablation of Adaptation Modules.}
Table~\ref{tab:Ablation Study of different settings.} and Fig. \ref{fig:Ablation of pseudo-label update mechanism.} evaluate the impact of different combinations of loss functions and the dynamic pseudo-label updating mechanism. To clearly assess the contribution of each module, we select the optimal model variant for each scenario: OmniSAM-L for indoor scenes and OmniSAM-B for outdoor scenes. For outdoor scenes, each proposed loss function positively contributes to segmentation performance, achieving an overall maximum improvement of $5.97\%$ in mIoU. In contrast, the prototypical adaptation shows limited effectiveness in indoor scenarios, primarily due to the negligible domain gap. Notably, the pseudo-labeling mechanism consistently proves effective, achieving a performance gain of $2.21\%$ when pseudo labels are dynamically updated.

\begin{table}[t!]
    \centering
    \renewcommand{\tabcolsep}{12pt}
    \resizebox{0.47\textwidth}{!}{
    \begin{tabular}{ccc|c|cccc}
        \toprule
        \multicolumn{3}{c|}{Loss Combinations} & \multirow{2}{*}{Update PL} & \multicolumn{2}{c}{CS13-DP13}\\
        \cmidrule(lr){1-3} \cmidrule(lr){5-6}
        $\mathcal{L}_{sup}$ & $\mathcal{L}_{ssl}$ & $\mathcal{L}_{fpa}$ & & mIoU & $\Delta$ \\
        \midrule
        \checkmark & \ding{55} & \ding{55}& \ding{55}& 56.49 & - \\
        \checkmark & \checkmark & \ding{55} & \ding{55}& 57.06 & +0.57 \\
        \checkmark & \checkmark & \checkmark & \ding{55}& 58.29 & +1.80  \\
        \checkmark & \checkmark & \ding{55}& \checkmark & 60.60 & +4.11  \\
        \checkmark & \checkmark & \checkmark & \checkmark & \textbf{62.46} & \textbf{+5.97} \\
        \bottomrule
        \toprule
        \multicolumn{3}{c|}{Loss Combinations} & \multirow{2}{*}{Update PL} & \multicolumn{2}{c}{SPin8-SPan8}\\
        \cmidrule(lr){1-3} \cmidrule(lr){5-6}
        $\mathcal{L}_{sup}$ & $\mathcal{L}_{ssl}$ & $\mathcal{L}_{fpa}$ & & mIoU & $\Delta$ \\
        \midrule
        \checkmark & \ding{55} & \ding{55}& \ding{55}& 76.85 & - \\
        \checkmark & \checkmark & \ding{55} & \ding{55}&  77.36 & +0.51 \\
        \checkmark & \checkmark & \checkmark & \ding{55}& 77.10 & +0.25 \\
        \checkmark & \checkmark & \checkmark & \checkmark & 77.79 & +0.94 \\
        \checkmark & \checkmark & \ding{55} & \checkmark & \textbf{79.06} & \textbf{+2.21} \\
        \bottomrule
    \end{tabular}
    }
    \vspace{-8pt}
    \caption{Ablation Study of different settings. (PL: Pseudo label)}
    \vspace{-16pt}
    \label{tab:Ablation Study of different settings.}
\end{table}

\begin{figure}[t!]
    \centering
    \includegraphics[width=0.47\textwidth]{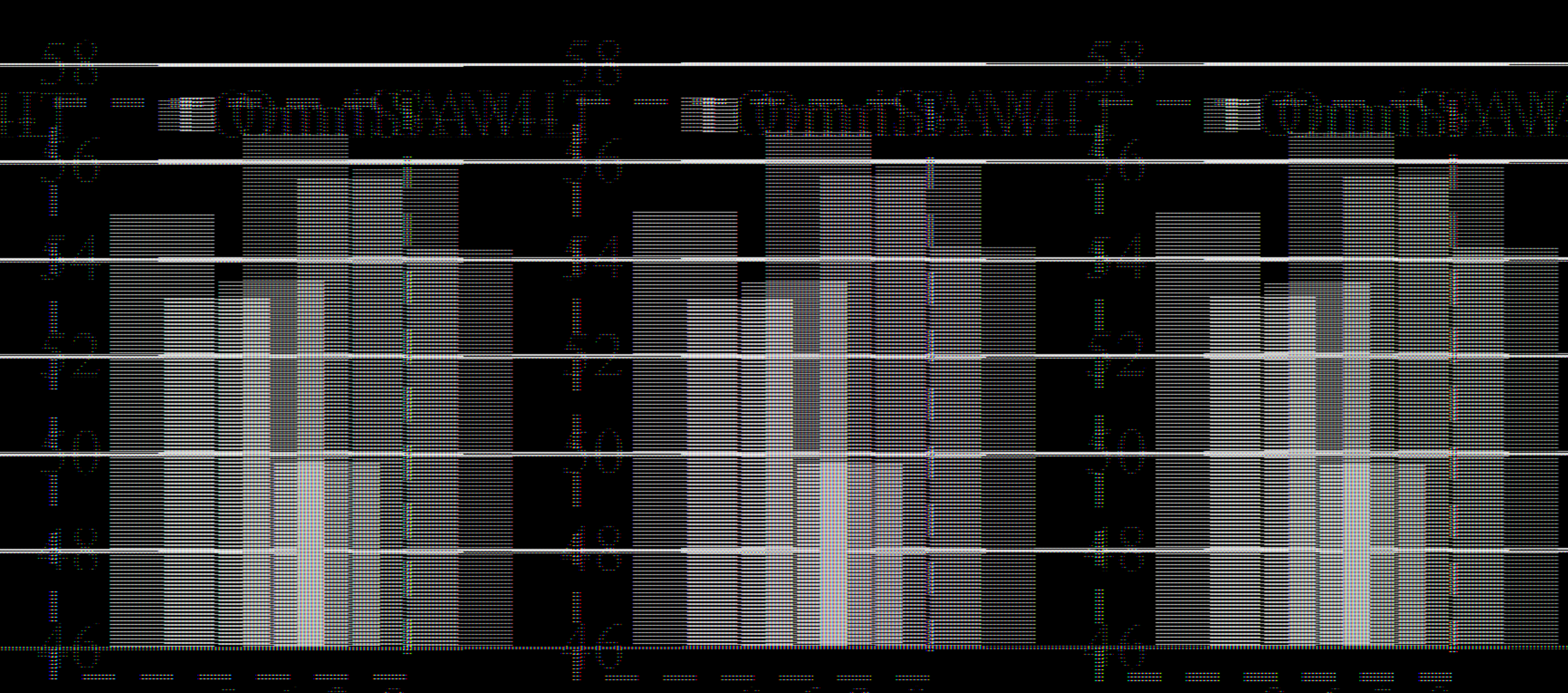}
    \vspace{-28pt}
    \caption{Ablation study of memory bank size on DensePASS.} 
    \vspace{-8pt}
    \label{fig:Ablation study of the size of memory bank.}
\end{figure}

\noindent \textbf{Ablation of Memory Bank Size.}
Next, we evaluate the impact of different memory bank sizes. Given that our sliding window mechanism extracts batches of 9 consecutive frames from the original image sequence, we investigate four memory bank sizes (0, 3, 6, and 9) to assess their effectiveness, as shown in Fig. \ref{fig:Ablation study of the size of memory bank.}. The results demonstrate that utilizing the largest memory bank size achieves the highest performance improvement.

\section{Conclusion}
In this paper, we presented our proposed OmniSAM for panoramic semantic segmentation. Leveraging the SAM2 pretrained image encoder and memory mechanism, we obtained a better source model though LoRA fine-tuning. For real world panoramic domain adaptation, we introduced a FPA module for patch-level prototypical adaptation to effectively align features across domains and improve robustness in distorted environments. Meanwhile, the dynamic pseudo-label updating strategy to iteratively improve the quality of target domain labels for self-supervised training. Experiments on real-world Pin2Pan and Syn2Real benchmarks show that our method outperforms all baseline models, achieving new state-of-the-art performance.

\noindent\textbf{Limitation and future work:} 
Our OmniSAM excels at real-world benchmarks, such as CS13-to-DP13 and SPin8-to-SPan8, while there is still room for improvement in SP13-to-DP13 scenario.
In the future, we plan to improve the performance in Syn2Real panoramic semantic segmentation scenarios.

\clearpage
\section{Acknowledgement}
This work was supported by the Guangdong Provincial Department of Education Project (Grant No.2024KQNCX028); CAAI-Ant Group Research Fund; Scientific Research Projects for the Higher-educational Institutions (Grant No.2024312096), Education Bureau of Guangzhou Municipality; Guangzhou-HKUST(GZ) Joint Funding Program (Grant No.2025A03J3957), Education Bureau of Guangzhou Municipality.
{
    \small
    \bibliographystyle{ieeenat_fullname}
    \bibliography{main}
}
\clearpage
\section*{Supplementary Material}
\renewcommand\thesubsection{\arabic{subsection}}
\setcounter{subsection}{0} 
\subsection{Supplementary Implementation Details}
Since the SAM2 backbone only accepts square-shaped images, we design two processing pipelines for source model training to ensure generalization across different cases: one for rectangular inputs and another for square-shaped inputs. 

\noindent\textbf{Rectangular Source Image.}
We apply a sliding window approach to segment the image into sequences of square patches, maintaining spatial continuity. These video-like sequences are then fed into the model to train the backbone, segmentation head, and memory modules. At each prediction step within an input sequence, the backbone feature integrates past memory embeddings through memory attention, allowing for more temporally consistent predictions. Subsequently, the current output mask and backbone feature are passed to the memory encoder, where the resulting embedding is stored in the memory bank for future reference. 

\noindent\textbf{Square Source Image.}
For inherently square-shaped inputs (\textit{Stanford2D3D-Pinhole}), treating them in the same manner as rectangular images would be unnatural. In this case, a trade-off is made by directly training the source model on these squared image-mask pairs without utilizing the memory mechanism, keeping the parameters of the memory attention and memory encoder frozen. Given that panoramic images are predominantly rectangular, the memory-related modules are fine-tuned on target domain data in the UDA stage to enhance feature alignment and improve segmentation performance in the panoramic domain.

\noindent\textbf{Training and Adaptation Details.}
During training, We introduce randomness to the window's sliding direction, including forward sliding (left-to-right) and reverse sliding (right-to-left). This helps model learn more cross-patch dependencies. For target domains in our domain adaptation, the SPan dataset contains \(4096\times2048\) images, which include inherent black regions brought by equirectangular projection. Hence we remove these redundant areas by cropping the images and downscale them to \(3072\times1024\). The corresponding sliding stride is \(256\). The DensePASS dataset contains \(2048\times400\) images, which are resized to \(3072\times1024\), with a corresponding sliding stride of \(256\). Besides, we sample 400 images from the target domain for pseudo-labels updating in each epoch.

\subsection{Supplementary Experimental Results}
\subsubsection{Trainable Parameters}
We evaluate the trainable parameters of our OmniSAM model by examining the impact of adding or removing the memory attention module. The parameter sizes for different variants of the OmniSAM model are presented in the Table \ref{tab:Comparison of trainable parameters for different variants of OmniSAM.}. Without memory attention, fine-tuning via LoRA-based adaptation affects only a small subset of the model, resulting in minimal trainable parameters across all variants. The smallest (OmniSAM-T) contains 0.36 MB, while the largest (OmniSAM-L) has 0.70 MB.
In contrast, the memory attention module contributes approximate 5.65 MB parameters. Specifically, OmniSAM-T increases from 0.36 MB to 6.01 MB, and OmniSAM-L grows from 0.70 MB to 6.35 MB. 
\begin{table}[ht]
    \centering

    \setlength{\tabcolsep}{16pt}
    
    \normalsize
    \resizebox{0.45\textwidth}{!}{
    \begin{tabular}{lc}
    \toprule
    Network & Trainable Param. (MB) \\
    \midrule
    OmniSAM-T w/o MA& 0.36 \\
    OmniSAM-S w/o MA& 0.39 \\
    OmniSAM-B w/o MA& 0.45 \\  
    OmniSAM-L w/o MA& 0.70 \\
    \midrule
    OmniSAM-T w/ MA& 6.01 \\
    OmniSAM-S w/ MA& 6.04 \\
    OmniSAM-B w/ MA& 6.10 \\  
    OmniSAM-L w/ MA& 6.35 \\
    \bottomrule
    \end{tabular}
    }
    \caption{Comparison of trainable parameters for different variants of OmniSAM.}
    \label{tab:Comparison of trainable parameters for different variants of OmniSAM.}
\end{table}
\subsubsection{Total Parameters and Computation Costs}
The total parameters and computation costs of our OmniSAM variants are presented in the Table \ref{tab:Comparison of total parameters for different variants of OmniSAM.}. The \textit{\textbf{Tiny}} and \textit{\textbf{Small}} variants exhibit competitive parameter counts, similar to state-of-the-art methods. While the \textit{\textbf{Large}} (209.2M) and the \textit{\textbf{Base}} (72.0M) variants significantly increase the model size, which may limit its real-time application. A trade-off in inference speed should be considered while choosing the model for specific task.
\begin{table}[ht]
    \centering
    \setlength{\tabcolsep}{16pt}
    \label{tab:results}
    
    \normalsize
    \resizebox{0.45\textwidth}{!}{
    \begin{tabular}{lcc}
    \toprule
    Network & Param. (M) & FLOPs (G) \\
    \midrule
    Trans4PASS+-S & 44.9 & 251.1 \\
    DATR-S & 25.8 & 139.2 \\
    360SFUDA++ & 28.7 & 148.0 \\ \midrule
    OmniSAM-T w/ MA & 32.0 & 118.8 \\
    OmniSAM-S w/ MA & 38.8 & 149.8 \\
    OmniSAM-B w/ MA & 72.0 & 280.8 \\  
    OmniSAM-L w/ MA & 209.2 & 828.0 \\
    \bottomrule
    \end{tabular}
    }
    \caption{Computatioal costs for networks with $1024^2$ resolution input.} 
    \label{tab:Comparison of total parameters for different variants of OmniSAM.}
\end{table}
\subsubsection{Ablation of $\lambda$.}
We conduct experiments on OmniSAM-B in the outdoor scenario to evaluate the impact of $\lambda$. As shown in Table \ref{tab:Ablation Study of lambda.}, OmniSAM-B achieves the highest score in mIoU while $\lambda=0.1$. 
\begin{table}[t!]
    \centering
    \renewcommand{\tabcolsep}{10pt}
    \resizebox{0.45\textwidth}{!}{
    \begin{tabular}{cccccccc}
        \toprule
        $\lambda$ & 0 & 0.01 & 0.1 & 0.2 & 0.5 & 1.0 \\
        \midrule
        mIoU & 56.49 & 61.64 & \textbf{62.46} & 61.59 & 61.44 & 60.90 \\
        $\Delta$ & - & +5.15 & \textbf{+5.97} & +5.10 & +4.95 & +4.41 \\
        \bottomrule
    \end{tabular}
    }
    \vspace{-8pt}
    \caption{Ablation Study of $\lambda$.}
    \vspace{-16pt}
    \label{tab:Ablation Study of lambda.}
\end{table}
\subsubsection{Ablation Study on Memory Mechanism}
Table \ref{tab:Per-class results of the Cityscapes13-to-DensePASS13 scenario (supp)} and Table \ref{tab:Per-class results of the Stanford2D3D pinhole-to-panoramic scenario (supp)} present additional class-specific results of our OmniSAM in the real-world scenario. The outcomes also serve as an ablation study on the memory mechanism of the model. 
\begin{table*}[t!]
    \centering
    \setlength{\tabcolsep}{3pt}

    \normalsize
    
    \resizebox{\textwidth}{!}{
    \begin{tabular}{llccccccccccccccc}
    \toprule
    {Method} & {Network} & {mIoU} & {Road} & {S.Walk} & {Build.} & {Wall} & {Fence} & {Pole} & {Tr.L} & {Tr.S} & {Veget.} & {Terrain} & {Sky} & {Person} & {Car} \\
    \midrule
    
    \multirow{8}{*}{Source-only} 
    & OmniSAM-T w/o MA& 49.79 & 71.91 & 26.12 & 86.42 & 36.34 & 39.73 & 36.29 & 8.27 & 13.04 & 79.47 & 20.92 & 94.16 & 59.10 & 75.53  \\ 
    & OmniSAM-S w/o MA& 53.98 & 75.28 & 36.88 & 87.69 & 42.41 & 40.89 & 38.45 & 19.19 & 14.03 & 80.08 & 27.48 & 94.36 & 63.59 & 81.38  \\
    & OmniSAM-B w/o MA& 55.03 & 76.69 & 38.98 & 88.60 & 42.56 & 48.07 & 40.18 & 20.43 & 14.79 & 81.04 & 29.38 & 94.83 & 60.73 & 79.15 \\
    & OmniSAM-L w/o MA & 54.23 & 75.70 & 36.29 & 87.23 & 41.29 & 47.53 & 38.98 & 20.52 & 14.29 & 81.18 & 28.75 & 94.54 & 57.16 & 81.56  \\ \cmidrule{2-16}
    & OmniSAM-T w/ MA & 54.99 & 75.48 & 36.30 & 88.60 & 39.74 & 41.35 & 42.29 & 20.32 & 24.57 & 78.35 & 32.43 & 94.23 & 62.5 & 78.65  \\
    & OmniSAM-S w/ MA & 55.40 & 76.86 & 36.92 & 88.64 & 39.89 & 41.31 & 42.70 & 25.11 & 20.19 & 78.14 & 32.76 & 94.68 & 61.9 & 81.11  \\
    & OmniSAM-B w/ MA & 56.49 & 74.41 & 43.58 & 87.65 & \textbf{44.80} & 46.91 & 45.17 & 16.24 & 22.24 & 80.15 & 32.41 & 94.86 & 65.31 & 80.67  \\
    & OmniSAM-L w/ MA & 56.61 & 77.52 & 43.45 & 88.50 & 36.51 & 51.65 & 38.12 & 20.96 & 21.70 & 81.48 & 31.65 & 94.62 & 70.88 & 78.95 \\
    \midrule
    
    \multirow{8}{*}{Ours} 
    & OmniSAM-T w/o MA& 53.73 & 79.03 & 42.19 & 86.09 & 28.28 & 45.95 & 35.19 & 10.20 & 20.53 & 79.41 & 35.49 & 94.40 & 63.81 & 77.88 \\ 
    & OmniSAM-S w/o MA& 57.03 & 78.99 & 49.57 & 88.77 & 38.48 & 47.47 & 38.77 & 21.84 & 15.81 & 81.12 & 39.32 & 94.72 & 65.36 & 81.22 \\
    & OmniSAM-B w/o MA& 59.34 & 81.69 & 53.87 & 89.33 & 39.74 & 50.84 & 41.98 & 20.54 & 21.50 & 81.71 & 44.63 & 95.06 & 68.13 & 82.34 \\
    & OmniSAM-L w/o MA & 59.02 & 83.49 & 56.14 & 88.29 & 34.29 & 52.39 & 38.81 & 23.97 & 19.83 & \textbf{82.52} & 44.84 & \textbf{95.26} & 61.97 & \textbf{85.43} \\ \cmidrule{2-16}
    & OmniSAM-T w/ MA& 59.01 & 79.95 & 45.78 & 88.03 & 39.74 & 47.99 & 42.68 & 26.69 & 29.55 & 78.00 & 42.98 & 94.56 & 68.23 & 82.98 \\
    & OmniSAM-S w/ MA& 60.23 & 80.76 & 46.36 & 89.79 &  44.46 & 48.68 &  \textbf{45.32}&  29.33 & 25.15 & 79.51 &  46.28 & 94.41 & 68.90 & 84.06   \\
    & OmniSAM-B w/ MA&  \textbf{62.46} &  \textbf{84.02} &  \textbf{56.23} &  89.93 & 44.01 &  54.54 & 44.50 & 25.19 &  \textbf{33.42} &  81.77 &  \textbf{49.16} & 94.69 & \textbf{71.64} &  82.89 \\  
    & OmniSAM-L w/ MA&  61.63 &  82.45 &  53.65 & \textbf{90.05} & 44.00 &  \textbf{54.75} & 43.36 &  \textbf{30.99} & 28.27 &  80.04 & 43.59 & 94.48 &  70.70 &  84.88 \\
    \bottomrule
    \end{tabular}
    }
    \caption{Per-class results of the Cityscapes13-to-DensePASS13 scenario (* denotes the baseline)}
    \label{tab:Per-class results of the Cityscapes13-to-DensePASS13 scenario (supp)}
\end{table*}

\begin{table*}[ht]
    \centering
    \setlength{\tabcolsep}{9pt}
    \resizebox{\textwidth}{!}{
    \begin{tabular}{llccccccccccccccc}
    \toprule
    {Method} & {Network} & {mIoU} & {Ceiling} & {Chair} & {Door} & {Floor} & {Sofa} & {Table} & {Wall} & {Window} \\
    \midrule
    
    \multirow{4}{*}{Source-only} 
    & OmniSAM-T w/o MA& 66.70 & 91.34 & 62.32 & 32.20 & 93.75 & 37.62 & 69.98 & 80.64 & 65.73  \\
    & OmniSAM-S w/o MA& 69.56 & 91.06 & 67.76 & 44.86 & 94.73 & 38.36 & 77.55 & 82.54 & 59.63 \\
    & OmniSAM-B w/o MA& 72.32 & 92.21 & 71.40 & 43.93 & 94.47 & 49.28 & 79.28 & 83.08 & 64.94 \\
    & OmniSAM-L w/o MA& 76.85 & \textbf{93.86} & \textbf{73.61} & 65.10 & 95.04 & \textbf{55.51} & \textbf{83.42} & 88.03 & 60.28 \\
    \midrule
    
    \multirow{8}{*}{Ours} 
    & OmniSAM-T w/o MA& 68.72 & 92.12 & 64.62 & 35.54 & 94.21 & 37.70 & 74.33 & 81.70 & 69.50 \\
    & OmniSAM-S w/o MA& 70.65 & 91.46 & 65.41 & 55.10 & 94.56 & 33.88 & 75.61 & 84.53 & 64.66 \\ 
    & OmniSAM-B w/o MA& 73.09 & 91.63 & 69.33 & 60.43 & 94.45 & 36.16 & 76.22 & 85.26 & 71.26 \\ 
    & OmniSAM-L w/o MA& 78.02 & 93.79 & 71.19 & \textbf{78.07} & \textbf{95.17} & 47.25 & 81.77 & \textbf{89.85} & 66.51 \\ \cmidrule{2-11}
    & OmniSAM-T w/ MA& 69.10 & 92.10 & 64.60 & 36.97 & 94.25 & 38.86 & 74.16 & 81.78 & 70.09 \\
    & OmniSAM-S w/ MA& 70.81 & 91.74 & 66.46 & 66.74 & 94.88 & 12.35 & 77.43 & 86.90 & 69.95 \\
    & OmniSAM-B w/ MA& 74.72 & 91.06 & 66.65 & 69.31 & 94.57 & 36.79 & 76.98 & 86.58 & \textbf{75.87} \\ 
    & OmniSAM-L w/ MA& \textbf{79.06} &93.25 & 72.12 & 77.97 & 95.00 & 52.08 & 81.82 & 89.62 & 70.58 \\
    \bottomrule
    \end{tabular}
    }
    \vspace{-8pt}
    \caption{Per-class results of the Stanford2D3D pinhole-to-panoramic scenario (* denotes the baseline).}
    \label{tab:Per-class results of the Stanford2D3D pinhole-to-panoramic scenario (supp)}
\end{table*}

\subsubsection{More Visualization Results}
\begin{figure*}[t!]
    \centering
    \includegraphics[width=0.99\textwidth]{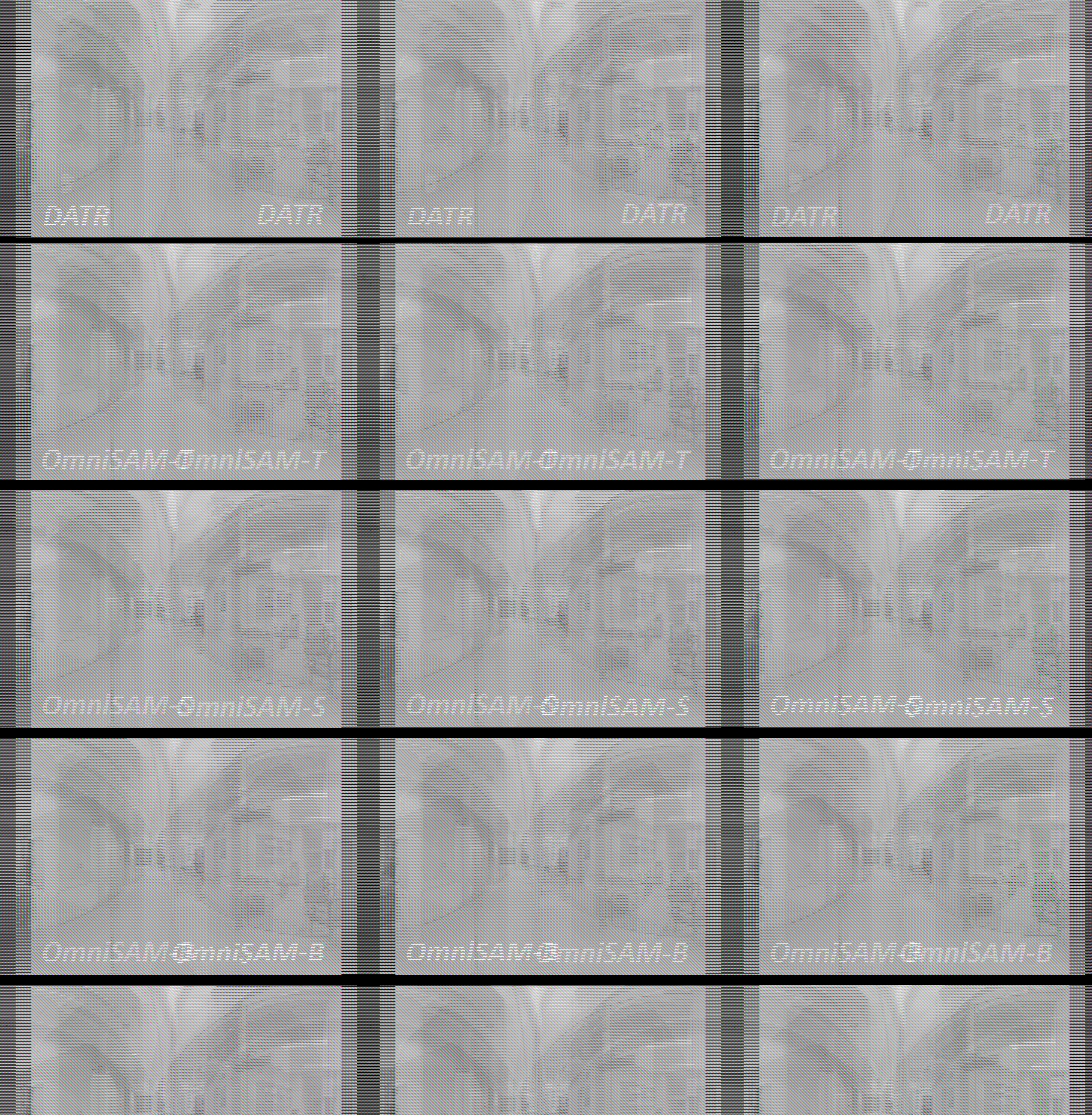}
    \caption{Visualizations on Stanford2D3D-Panoramic dataset for different variants of OmniSAM.}
    \label{fig:Visualizations on Stanford2D3D-Panoramic dataset for different variants of OmniSAM.}
\end{figure*}

\begin{figure*}
  \centering
  \begin{subfigure}{0.49\linewidth}
    \includegraphics[width=\linewidth]{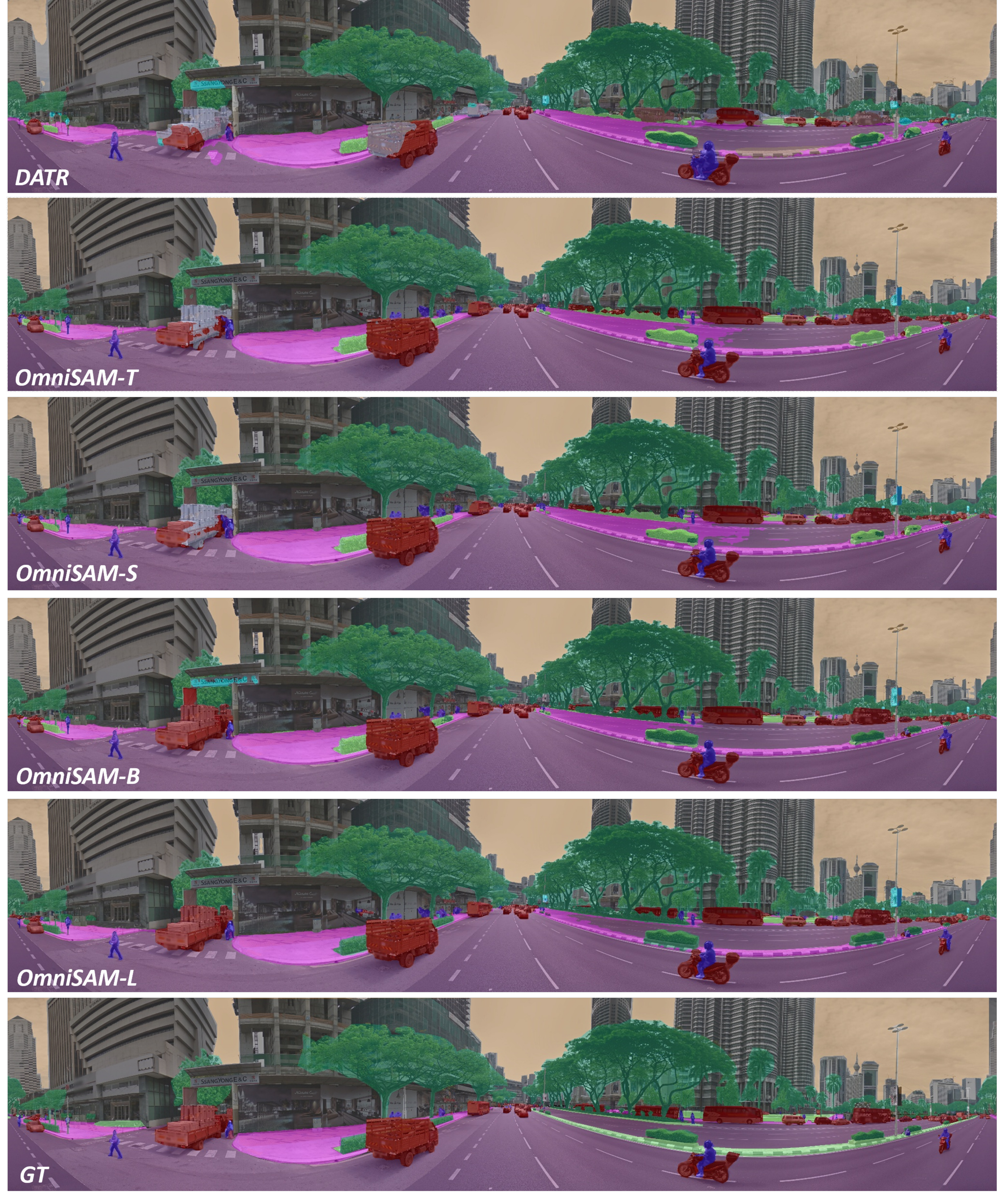}
    \caption{}
    \label{fig:short-a}
  \end{subfigure}
  \hfill
  \begin{subfigure}{0.49\linewidth}
    \includegraphics[width=\linewidth]{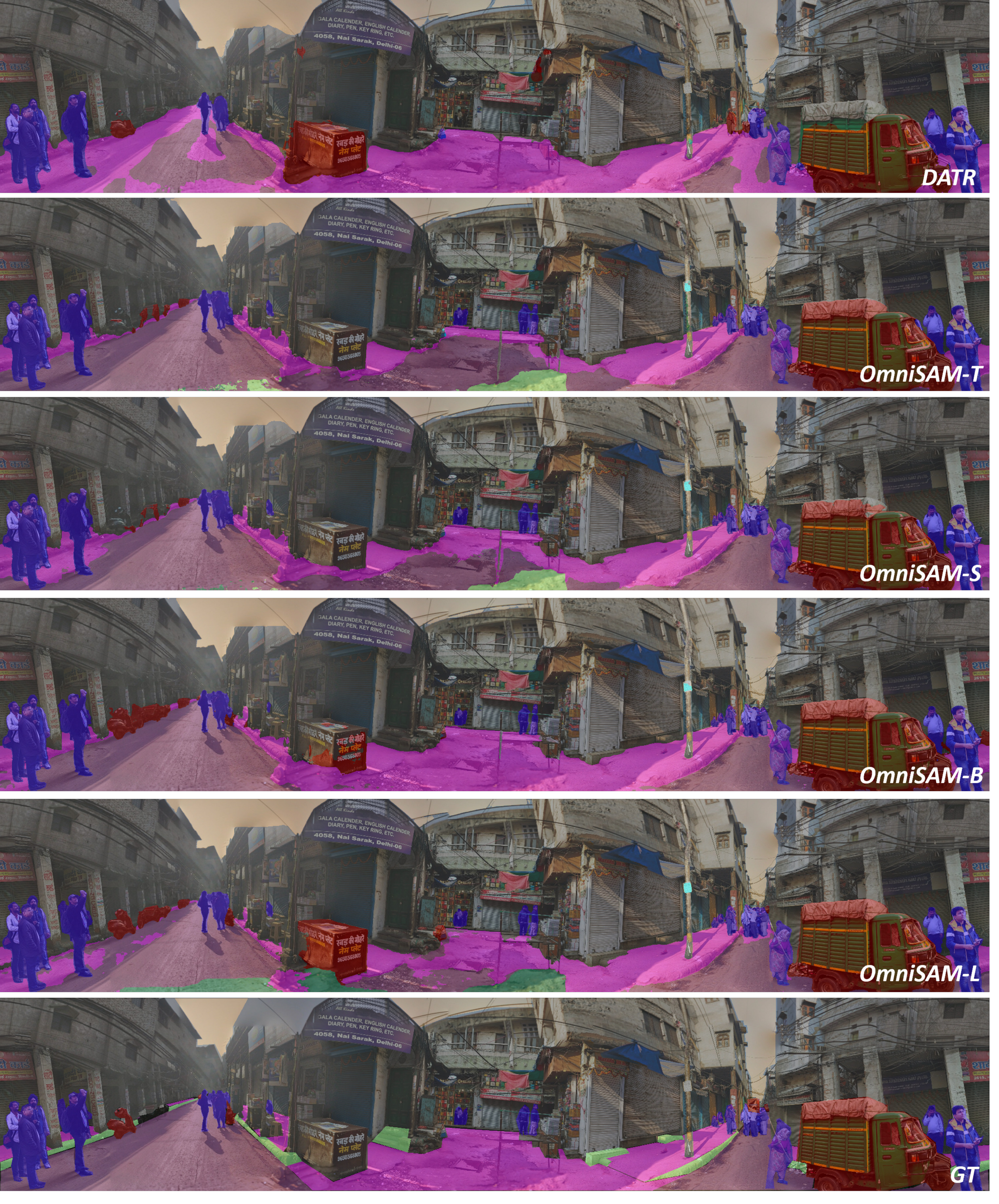}
    \caption{}
    \label{fig:short-b}
  \end{subfigure}

    \begin{subfigure}{0.49\linewidth}
    \includegraphics[width=\linewidth]{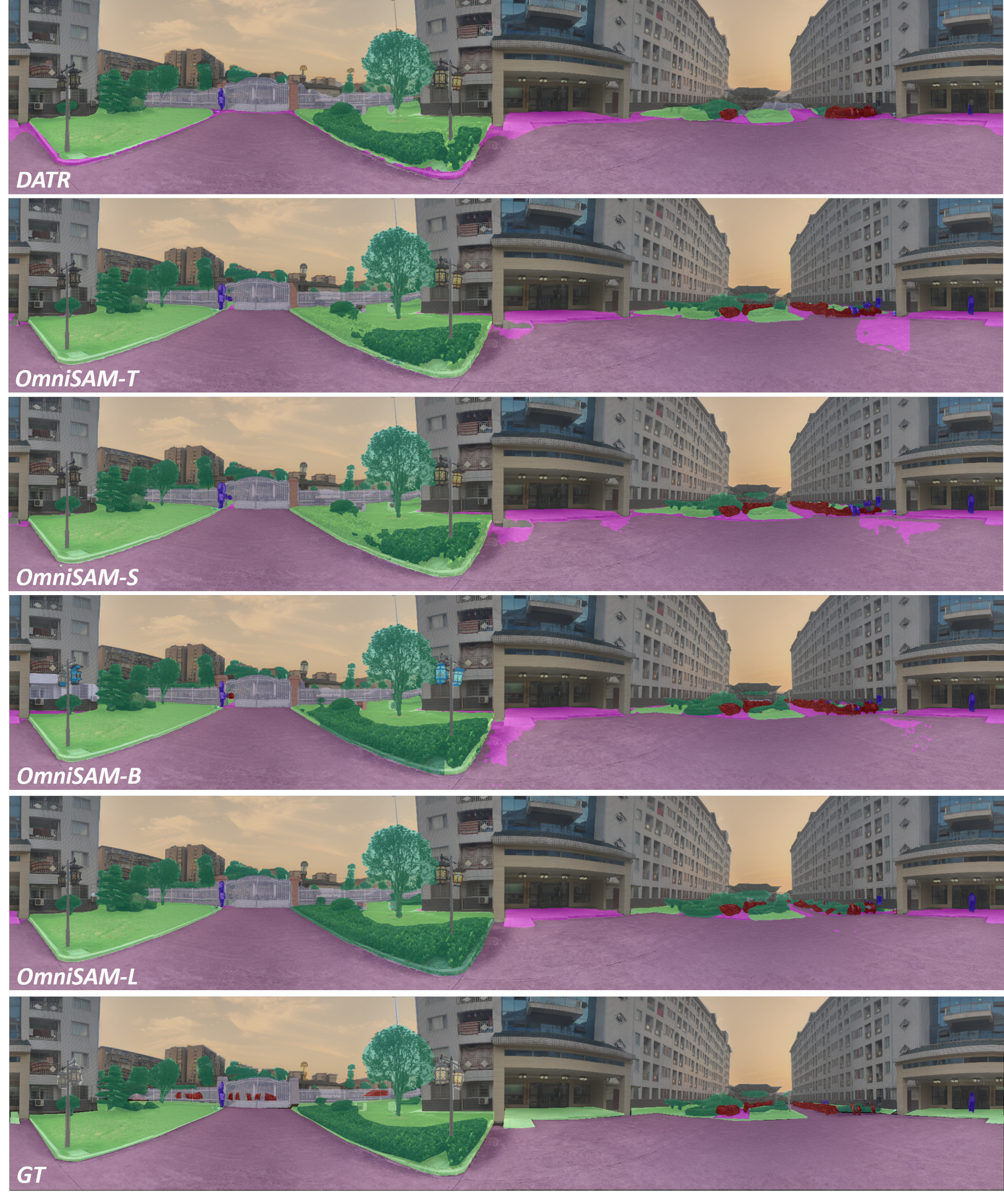}
    \caption{}
    \label{fig:short-c}
  \end{subfigure}
  \hfill
  \begin{subfigure}{0.49\linewidth}
    \includegraphics[width=\linewidth]{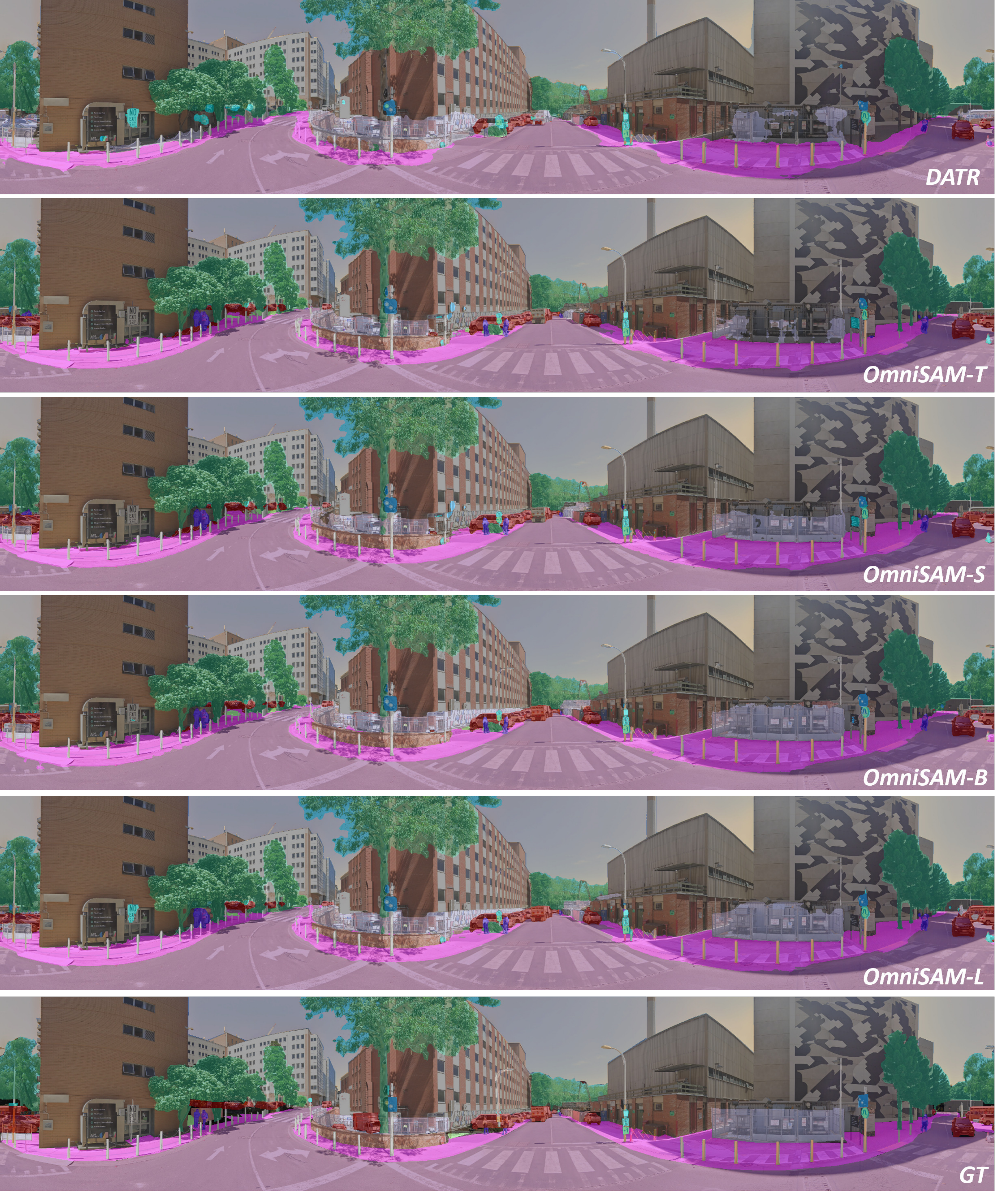}
    \caption{}
    \label{fig:short-d}
  \end{subfigure}
    \caption{Visualizations on DensePASS dataset dataset for different variants of OmniSAM.}
    \label{fig:Visualizations on DensePASS dataset dataset for different variants of OmniSAM.}
\end{figure*}

\begin{figure*}[t!]
    \centering
    \includegraphics[width=0.99\textwidth]{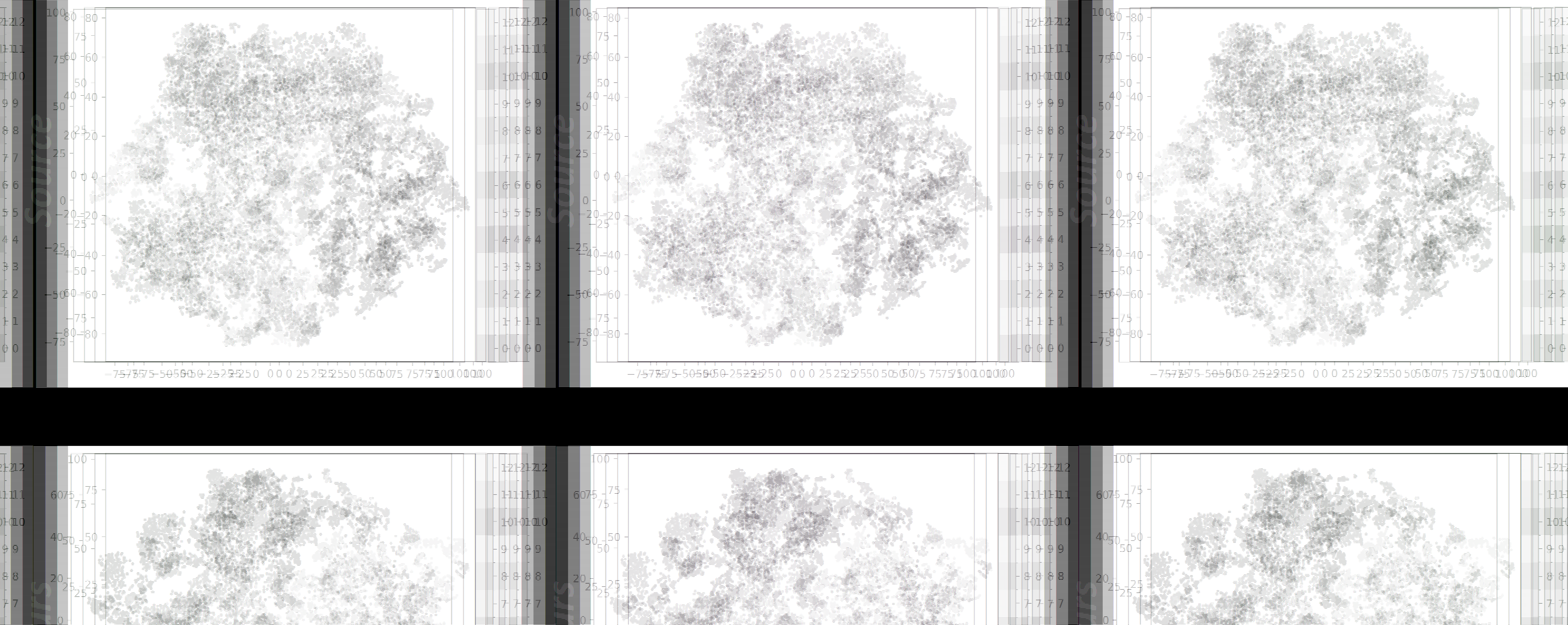}
    \caption{t-SNE Visualizations on DensePASS of Cityscapes-to-DensePASS.}
    \label{fig:TSNE Visualizations on DensePASS of Cityscapes-to-DensePASS.}
\end{figure*}

Fig.~\ref{fig:Visualizations on Stanford2D3D-Panoramic dataset for different variants of OmniSAM.} and Fig.~\ref{fig:Visualizations on DensePASS dataset dataset for different variants of OmniSAM.} present segmentation results from our proposed model and baseline methods~\cite{zheng2024360sfuda++,zheng2023look} across various settings. Additionally, the t-SNE visualization shown in Fig.~\ref{fig:TSNE Visualizations on DensePASS of Cityscapes-to-DensePASS.} further highlights the effectiveness of our adaptation approach by illustrating distinct and informative feature representations for each semantic category.

\end{document}